\documentclass[10pt,twocolumn,letterpaper]{article}

\usepackage{cvpr}              

\definecolor{Ours}{HTML}{EAFAD5}
\definecolor{lightgray}{HTML}{D3D3D3}

\definecolor{cvprblue}{rgb}{0.21,0.49,0.74}
\usepackage[pagebackref,breaklinks,colorlinks,allcolors=cvprblue]{hyperref}
\usepackage{kotex}
\usepackage{multirow}
\usepackage{algorithm}
\usepackage{amsmath}
\usepackage{algpseudocode}
\usepackage{subcaption} 
\usepackage[capitalize]{cleveref}
\usepackage{adjustbox}
\crefname{section}{Sec.}{Secs.}
\Crefname{section}{Section}{Sections}
\Crefname{table}{Table}{Tables}
\crefname{table}{Tab.}{Tabs.}

\newlength\savewidth

\newcolumntype{x}[1]{>{\centering\arraybackslash}p{#1pt}}
\newcolumntype{y}[1]{>{\raggedright\arraybackslash}p{#1pt}}
\newcolumntype{z}[1]{>{\raggedleft\arraybackslash}p{#1pt}}

\def\paperID{10935} 
\def\confName{CVPR}
\def\confYear{2025}

\title{Parameter Efficient Mamba Tuning \\ via Projector-targeted Diagonal-centric Linear Transformation}

\author{Seokil Ham \and Hee-Seon Kim \and Sangmin Woo \and Changick Kim \and
Korea Advanced Institute of Science and Technology (KAIST)\\
Daejeon, South Korea\\
{\tt\small \{gkatjrdlf, hskim98, smwoo95, changick\}@kaist.ac.kr}
}

\begin{document}
\maketitle
\begin{abstract}

Despite the growing interest in Mamba architecture as a potential replacement for Transformer architecture, parameter-efficient fine-tuning (PEFT) approaches for Mamba remain largely unexplored. 
In our study, we introduce two key insights-driven strategies for PEFT in Mamba architecture: (1) While state-space models (SSMs) have been regarded as the cornerstone of Mamba architecture, then expected to play a primary role in transfer learning, our findings reveal that Projectors---not SSMs---are the predominant contributors to transfer learning. (2) Based on our observation, 
we propose a novel PEFT method specialized to Mamba architecture: \textbf{Pro}jector-targeted \textbf{Dia}gonal-centric \textbf{L}inear Transformation (\textbf{ProDiaL}). ProDiaL focuses on optimizing only the pretrained Projectors for new tasks through diagonal-centric linear transformation matrices, without directly fine-tuning the Projector weights. This targeted approach allows efficient task adaptation, utilizing less than 1\% of the total parameters, and exhibits strong performance across both vision and language Mamba models, highlighting its versatility and effectiveness.

\end{abstract}
    
\section{Introduction}
\label{sec:intro}

Recently, Mamba architecture~\cite{gu2023mamba,dao2024transformers,zhu2024vision,
liu2024vmamba} has been garnered attention as a promising alternative to the Transformer architecture~\cite{vaswani2017attention}. While Transformers have become the foundational architecture for deep learning across various fields, they suffer from a significant limitation: their inference time increases quadratically with the length of input tokens. To overcome this limitation, Mamba architecture introduces a selective mechanism within state-space models (SSMs)~\cite{gu2021efficiently,fu2022hungry} and hardware-aware operations, allowing dynamic and linear computation with respect to input size.

\begin{figure}[t!]
\centering
    \includegraphics[width=\linewidth,]{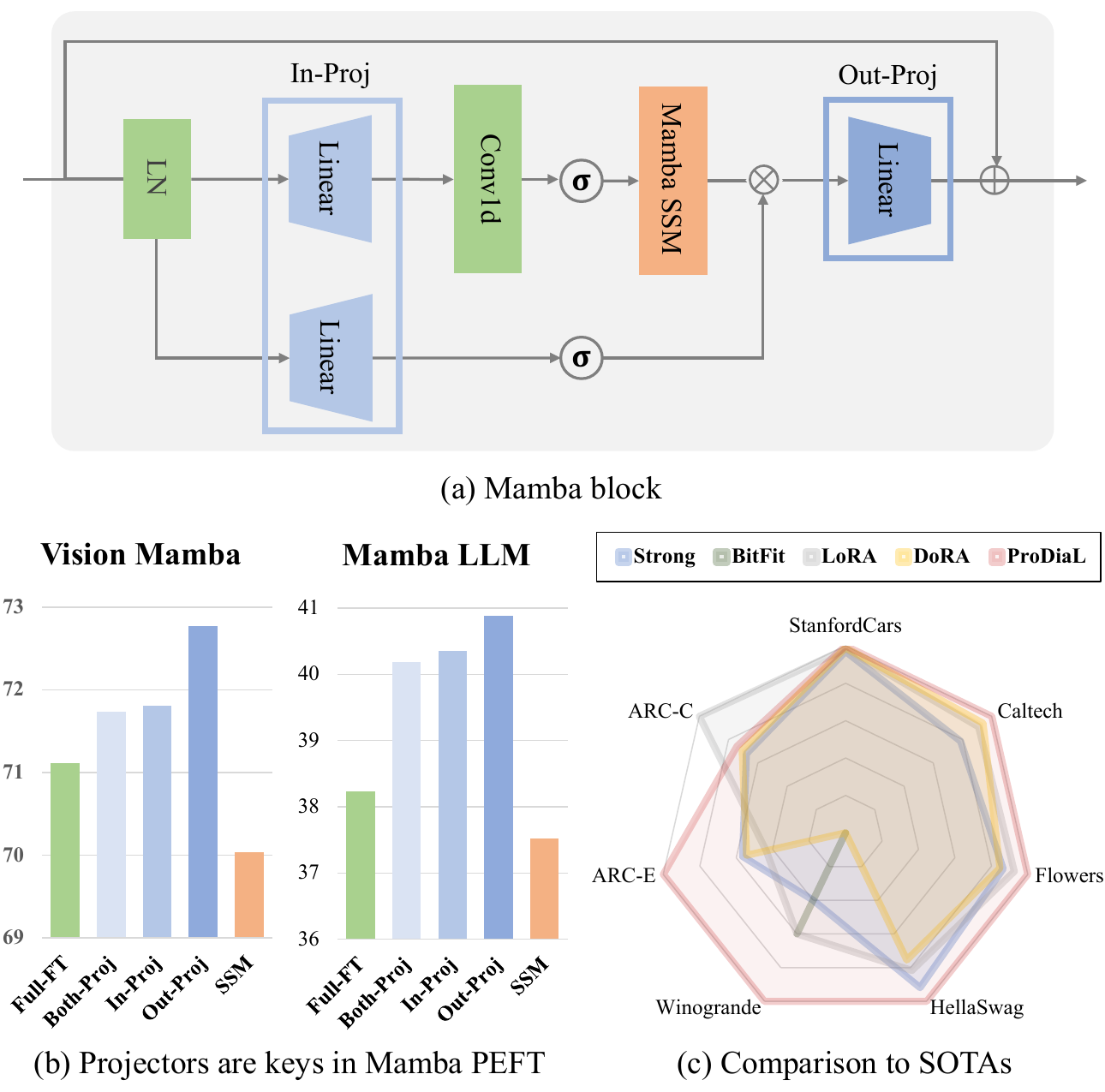}
    \vspace{-0.3cm}   
    \caption{
\textbf{ Overview of Mamba Architecture and Performance Comparison.} (a) The Mamba block structure, illustrating key components including the Input-Projector (In-Proj), Output-Projector (Out-Proj), and State-Space Model (SSM). (b) Performance analysis in Fine-Tuning for Vision Mamba and Mamba LLM, showing that projectors are essential for effective downstream task performance. (c) The radar chart illustrates the relative performance of our proposed method (ProDiaL) compared to other leading PEFT methods (Strong~\cite{halloran2024mamba}, BitFit~\cite{zaken2021bitfit}, LoRA~\cite{hu2021lora}, DoRA~\cite{liu2024dora}) across multiple benchmarks, demonstrating powerful performance in both vision and language tasks.
}
\label{fig:mamba_teasor}
\end{figure}

As an alternative to Transformers, Mamba architecture has become widely used in large-scale models~\cite{gu2023mamba,dao2024transformers,ren2024samba,glorioso2024zamba,lee2024meteor,teng2024dim,fei2024dimba} due to its computational efficiency. However, despite this efficiency, large-scale models based on Mamba still face significant computational and memory costs during full fine-tuning.
While existing parameter-efficient fine-tuning (PEFT) methods~\cite{chen2022adaptformer,jia2022visual,li2021prefix,zhang2020side,zaken2021bitfit,hu2021lora} are designed primarily for Transformer-based architectures, leaving PEFT methods  tailored to Mamba largely unexplored.

In this work,  we first investigate the Mamba architecture to determine the components best suited for PEFT. Among the key operations of the Mamba architecture—--SSM, 1D-Convolution, Embedding, and Linear Projectors—--previous study~\cite{halloran2024mamba} has focused on fine-tuning the SSM, assuming it to be critical for downstream task adaptation. However, our findings reveal that the Projectors, rather than the SSM, play a crucial role in  adapting knowledge for downstream tasks. To the best of our knowledge, this is the first study to conduct an in-depth exploration of the Projectors in the Mamba architecture from a transfer learning perspective.

To efficiently and effectively fine-tune the Projectors for downstream tasks, we propose a novel PEFT method targeting the Projectors in the Mamba architecture, called \textbf{Pro}jector-targeted \textbf{Dia}gonal-centric \textbf{L}inear Transformation (\textbf{ProDiaL}). 
Our ProDiaL builds upon our two key observations: (1) the pretrained Projector weight $W$ can be indirectly fine-tuned by training the linear transformation matrix $T$, where $WT$ approximates a fine-tuned Projector weight $W'$, and (2) $T$ is a nearly an identity matrix with strong diagonal values and minimal off-diagonal components.
Inspired by these insights, ProDiaL adapts pretrained Projectors to downstream tasks by decomposing the transformation matrix $T$ into diagonal and off-diagonal matrices: $W' = WD + \epsilon$, where $D$ is a diagonal matrix, and $\epsilon$ represents off-diagonal matrix. Furthermore, to enable more expressive transformations, we use a block-diagonal structure for $D$, and adopt LoRA~\cite{hu2021lora} for $\epsilon$, which matches $W$ in dimension. As a result, our ProDiaL trains only a block-diagonal matrix for $D$ and two small matrices for $\epsilon$, with the number of parameters controlled by low-rank factors. This design achieves parameter efficiency while preserving flexibility. 

Our proposed ProDiaL is specifically designed based on the analysis of Mamba Projectors, making it compatible with all models built upon Mamba architecture such as Mamba-based Large Language Models (LLMs)~\cite{gu2023mamba} and Vision models~\cite{zhu2024vision}. Our experiment results demonstrate that fine-tuning the Projectors is crucial for adapting the pretrained model to downstream tasks, and our proposed ProDiaL yields superior performance in downstream tasks while training less than 1\% of the total model parameters, highlighting its efficiency and effectiveness on both Mamba-based vision and language models.

\subsection*{Our Contributions}
\begin{itemize}
    \item By investigating the impact of each component in Mamba for transfer learning, we discover that Mamba’s PEFT primarily benefits from fine-tuning the Projectors.\\
    \vspace{-3mm}
    \item With our analysis that fine-tuned Projectors in Mamba can be approximated as a near-diagonal linear transformation of pretrained Projectors, we propose ProDiaL.\\
    \vspace{-3mm}
    \item Experiments on both Vision Mamba and Mamba LLM demonstrate that applying not only our proposed ProDiaL method but also existing PEFT methods to the Projectors significantly outperforms targeting other components.\\
\end{itemize}
\vspace{-3mm}

\section{Related work}
\label{sec:relwork}

\subsection{State Space Models}

\textit{State Space Models} (SSMs)~\cite{kalman1960new} are dynamic systems that represent the relationships between inputs, hidden states, and outputs over time.
A notable example, \textbf{Structured State Space Model (S4})~\cite{gu2021efficiently,gu2021combining,gupta2022diagonal,smith2022simplified,nguyen2022s4nd,islam2022long} is designed to handle long-range sequences with Linear Time Invariance (LTI) system, which ensures consistent outputs for identical inputs regardless of their temporal positions in a sequence. A key advantage of S4 is that its computational cost scales linearly with sequence length.
However, the fixed internal state transition matrix over time restricts the model’s flexibility in adjusting to changing content, limiting its effectiveness for tasks that require context-based reasoning.
As a result, despite of the quadratic computational cost with respect to the sequence length, Transformers are preferred for processing sequential data~\cite{vaswani2017attention,touvron2023llama,brown2020language,radford2023robust}. 

To address these limitations of SSMs, \textbf{Mamba}~\cite{gu2023mamba,dao2024transformers} introduces a selective mechanism and hardware-aware operation to overcome the quadratic computational costs of Transformers.
This enables Mamba to support context-aware reasoning with linear computational cost, extending its applications to various sequential data tasks including language and speech tasks~\cite{gu2023mamba,ren2024samba,glorioso2024zamba}. 
Similar to Transformers, Mamba is also being applied in the vision domain. \textbf{Vision Mamba} models~\cite{zhu2024vision,li2024mamba,park2025videomamba,liu2024vmamba,teng2024dim,fei2024dimba} adopt bidirectional scanning methods to effectively represent 2-D spatial information of images as 1-D sequences. This feature extracting strategy has proven effective, being adopted not only in simple vision tasks such as classification~\cite{zhu2024vision,liu2024vmamba,li2024mamba,park2025videomamba} but also in more complex vision tasks like image generation tasks~\cite{teng2024dim,fei2024dimba}.

Despite the success of these Mamba models and the growing scale of Mamba models, PEFT of Mamba architectures for downstream tasks remains largely unexplored. Therefore, in this work, we uncover the core component in Mamba architecture relevant to PEFT and propose a novel PEFT method tailored to Mamba, called ProDiaL, based on new insights from our analysis.

\subsection{Parameter Efficient Fine-Tuning}
Fine-tuning large models often demands significant computational and memory resources, especially when working with limited data, which might increase the risk of overfitting.
To address these challenges, \textit{parameter-efficient fine-tuning} (PEFT) methods have emerged, particularly targeting Transformer-based architectures.
Key PEFT techniques include:
\textbf{(1) Adapters} integrate learnable modules within pre-trained models, allowing them to train separately from the model’s primary frozen weights.
Adapter variations include serial configurations (\eg, Serial Adapter~\cite{houlsby2019parameter}) and parallel structures (\eg, AdaptFormer~\cite{chen2022adaptformer}, ControlNet~\cite{zhang2023adding}).
\textbf{(2) Prompt Tuning} uses trainable embeddings that guide downstream learning in visual tasks, optimizing model adaptation for vision applications (\eg, Visual Prompt Tuning~\cite{jia2022visual}, LLaMA-Adapter~\cite{zhang2023llama}, DMP~\cite{ham2024diffusion}).
\textbf{(3) Subset Fine-tuning} modifies specific model parameters, such as bias terms (\eg, BitFit~\cite{zaken2021bitfit}, DiffFit~\cite{xie2023difffit}) or selected key-value weights (\eg, Custom Diffusion~\cite{kumari2023multi}) within Transformer attention layers, thereby reducing training overhead.
\textbf{(4) Low-Rank Adaptation (LoRA)}~\cite{hu2021lora} incorporates low-rank matrices for fine-tuning, preserving the original model weights while facilitating effective downstream learning.
This method is particularly useful for conserving model integrity within Transformer architectures (\eg, VeRA~\cite{kopiczko2023vera}, DoRA~\cite{liu2024dora}, QLoRA~\cite{dettmers2024qlora}, MTLoRA~\cite{agiza2024mtlora}).

Despite the success of these PEFT methods with Transformer models~\cite{vaswani2017attention}, their application to Mamba architectures~\cite{gu2023mamba,dao2024transformers,zhu2024vision,liu2024vmamba} remains largely unexplored.
Recently, Halloran~\etal~\cite{halloran2024mamba} examined the use of PEFT techniques, specifically LoRA, within Mamba SSMs.
Unlike previous works, we reveal that PEFT in Mamba architectures is more effective when applied to Projectors rather than SSMs.
Building on this insight, we introduce ProDiaL, a novel PEFT method focused on efficient adaptation of Projectors, specialized to Mamba architecture.

\section{Preliminary}
To address the limitations inherent in transformer architectures---specifically, the quadratic increase in computational cost with respect to the number of input tokens---the Mamba architecture is introduced as a novel solution centered around the SSM.

Initially, the SSM is a LTI system that maps an input sequence $x(t)$ to a hidden state $h(t)$ and predicts the output $y(t)$ by leveraging both $x(t)$ and $h(t)$. 
The process is defined by the following system of Ordinary Differential Equations (ODEs):

\begin{equation}
\label{eq:ssm_continuous}
\begin{split}
h'(t)={A}h(t)+{B}x(t),\\
y(t)={C}h(t)+{D}x(t),
\end{split}
\end{equation}

where $x(t) \in \mathbb{R}$ is the continuous input sequence, $h(t) \in \mathbb{R}^{N}$ is the hidden state, and $y(t) \in \mathbb{R}$ is the  output sequence. While this equation addresses continuous signals, discretization is required for processing discrete signals. In Mamba, the Zero-Order Hold technique is used to discretize parameters $A$ and $B$, utilizing a step size parameter $\Delta$ to derive the discrete equivalents:
\begin{equation}
\label{eq:discretization}
\begin{split}
&\bar{A} = \exp(\Delta A),\\
&\bar{B} = (\exp(\Delta A) - I)(\Delta A)^{-1}B ,\\ 
&\bar{C} = C.\\
\end{split}
\end{equation}
With these discretized parameters $\bar{A}$, $\bar{B}$, and $\bar{C}$, 
the SSM is reformulated to address discrete signals as follows:
\begin{equation}
\label{eq:ssm_discrete}
\begin{split}
&h_n = \bar{A}h_{n-1} + \bar{B}x_n, \\ 
&y_n = \bar{C}h_n,\\
\end{split}
\end{equation}
where $n$ indexes the input sequence for the discrete signal. 
In contrast to the continuous formulation in~\cref{eq:ssm_continuous}, this discretized ODE explicitly separates current and previous states, with the hidden state $h_n$ computed from the current input $x_n$ and the previous state $h_{n-1}$.

The architecture of a Mamba block is a combination of the SSM-based model~\cite{fu2022hungry} and a gated MLP~\cite{liu2021pay}. Therefore, in addition to the SSM, Mamba blocks also include two Projectors and a 1D Convolution layer, as illustrated in ~\cref{fig:mamba_teasor}, representing the original Mamba block.

\section{Methodology}

Current PEFT methods are primarily designed for Transformer-based architectures. However, despite Mamba architecture is widely used in large-scale models such as LLMs or Diffusion models, it remains underexplored in terms of PEFT techniques.
In this section, we analyze the impact of each component in Mamba blocks on downstream task performance by fine-tuning various combinations of these components and propose a novel PEFT method specialized to Mamba architecture.

\subsection{Targeting Projectors for PEFT} \label{sec:targeting_projectors}
In Transformer-based models, PEFT methods typically target the attention module as the primary component for adapting to downstream tasks~\cite{hu2021lora, agiza2024mtlora, liu2024dora, kopiczko2023vera, dettmers2024qlora}. Following this prior knowledge, a natural approach for PEFT in  Mamba might be to fine-tune the SSM, which plays a core role similar to the attention mechanism in Transformers. 
However, the effectiveness of fine-tuning the SSM in the Mamba architecture has not actually been explored yet.

To address this, we first identify key components for downstream task learning in the Mamba architecture. Based on \cite{halloran2024mamba}, which fine-tuned parameters including $W_x$ (responsible for determining $B$, $C$, and $\Delta$ in the SSM), two Projectors, and Embeddings, we selectively fine-tune or exclude these components to assess their individual contributions to downstream performance.
The results, as shown in~\cref{table:observation vim,table:observation LLM}, highlight the following points:

\begin{table}[!t]
\centering
\begin{tabular}{lcc}
\hline
Target Components & \# of parameters & Accuracy \\ \hline
Full-FT                         & 6.975M   & 71.11  \\
$W_x$, Both-Proj~\cite{halloran2024mamba}             & 6.196M   & 71.39 \\
Both-Proj                            & 5.328M   & 71.74 \\
$W_x$, In-Proj                 & 4.427M   & 70.89  \\
In-Proj                        & 3.558M   & 71.81 \\
$W_x$, Out-Proj                & 2.657M   & 73.00  \\
Out-Proj                       & 1.789M   & 72.77   \\ 
SSM                             & 1.441M   & 70.04  \\ \hline
\end{tabular}
    \vspace{-2mm}
\caption{\textbf{Comparison of downstream task performance (\%) by fine-tuning various components in Vision Mamba model.} Notably, fine-tuning only projectors yield a high accuracy compared to targeting other components.}
\label{table:observation vim}
    \vspace{-2mm}
\end{table}

\begin{table}[!t]
\centering
\small
\begin{tabular}{lcc}
\hline
Target Components & \# of parameters & Accuracy \\ \hline
Full-FT                         & 130M   & 38.23  \\
$W_x$, Both-Proj, Embed~\cite{halloran2024mamba}     & 126M   & 38.68 \\
$W_x$, Both-Proj                     & 87.9M  & 39.53  \\ 
Both-Proj, Embed                     & 124M   & 39.06 \\
Both-Proj                            & 84.9M  & 40.18 \\
In-Proj                        & 56.6M  & 40.36 \\
Out-Proj                       & 28.3M  & 40.89   \\ 
$W_x$, In-Proj                 & 59.6M  & 39.76  \\
$W_x$, Out-Proj                & 31.3M  & 40.79  \\
SSM                             & 5.38M  & 37.52  \\
Embed                           & 38.6M  & 34.39  \\  \hline
\end{tabular}
    \vspace{-2mm}
\caption{\textbf{Downstream task performance(\%) results for Mamba LLM}, showing similar trends to the Vision Mamba model.}
\label{table:observation LLM}
	\vspace{-0.3cm}   
\end{table}

(1) \textbf{Projector Dominance}: Projectors play a more significant role in learning downstream task knowledge than $W_x$, even though $W_x$ directly affects the most of elements ($B$, $C$, and $\Delta$) in the SSM module.

(2) \textbf{Input vs. Output Projectors}: Fine-tuning either the Input Projector (a Projector before SSM) or the Output Projector (a Projector after SSM) yields performance comparable to fine-tuning both Projectors together.

(3) \textbf{Embedding Limitations}: Embedding parameters, as used in \cite{halloran2024mamba}, hinder the learning of downstream task knowledge, ultimately limiting performance.

These findings are consistent across both the Mamba LLMs~\cite{gu2023mamba} and Vision Mamba models~\cite{zhu2024vision}. Given these observations, one might question whether the higher downstream task performance of Projectors than other components is merely due to their larger number of parameters. However, as demonstrated in~\cref{table:main_results}, high downstream task performance preserves even with a reduced number of learnable parameters, suggesting that the effectiveness of Projectors is not solely attributable to parameter size.
However, comparing the performance of SSM with 5.38M parameters ($37.52\%$) to that of Both-Projectors using LoRA with 2.36M parameters ($38.33\%$) in~\cref{table:main_results}, high downstream task performance preserves even with a reduced number of learnable parameters. This suggests that the effectiveness of Projectors is not solely due to parameter size.

\begin{figure}[!t]
    \centering
    \begin{subfigure}{0.62\linewidth}
        \centering
        \includegraphics[width=\linewidth]{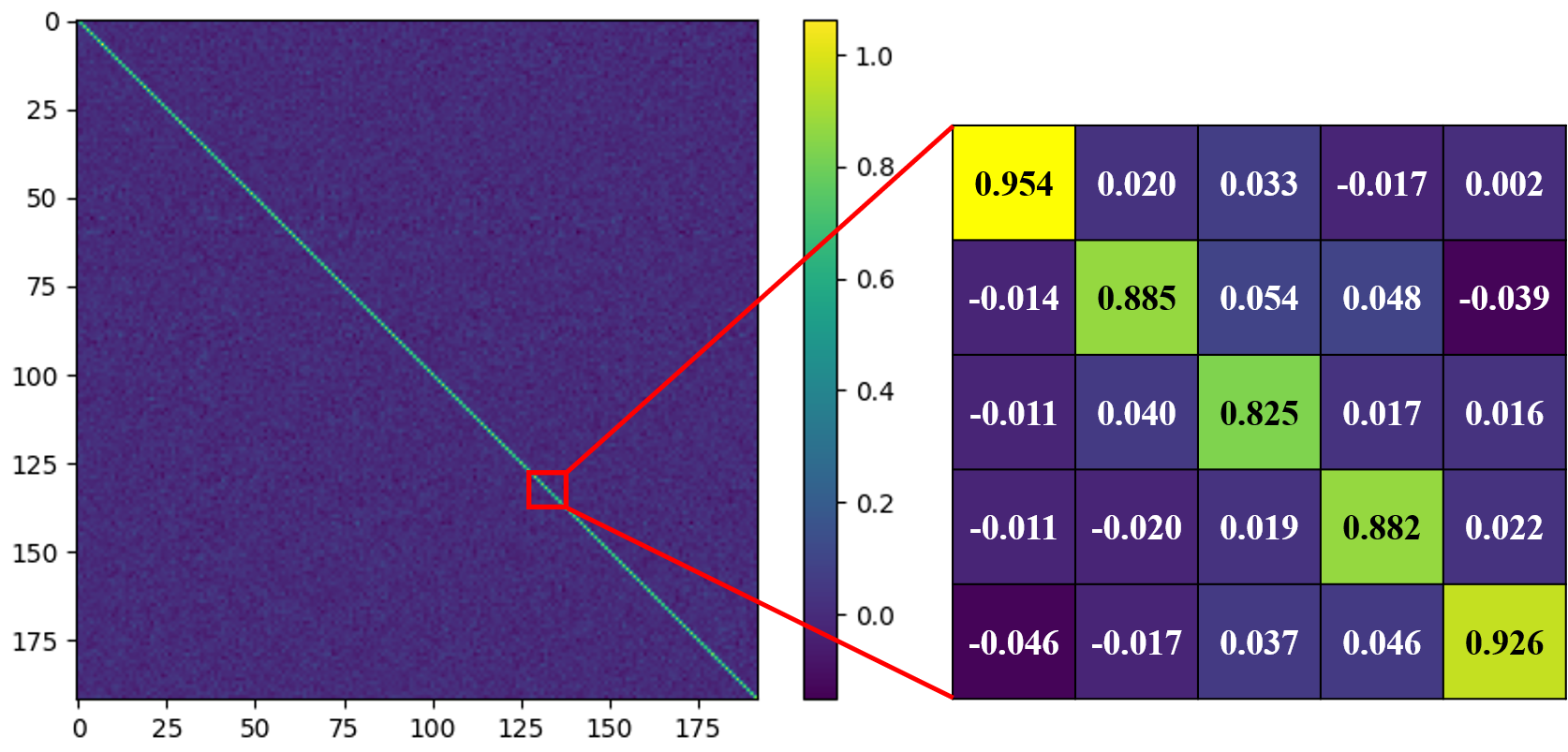}
        \caption{Visualization of the matrix $T_{det}$.}
        \label{fig:diagonal_analysis}
    \end{subfigure}%
    \hfill
    \begin{subfigure}{0.37\linewidth}
        \centering
        \includegraphics[width=\linewidth]{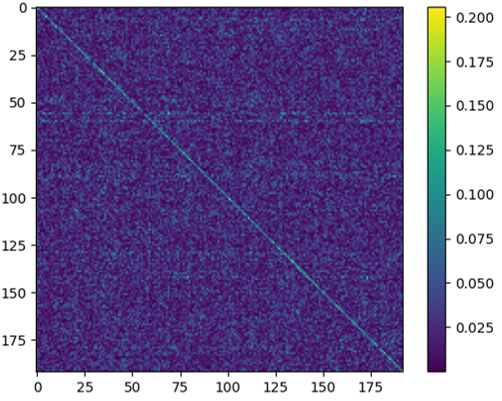}
        \caption{Changes after training.}
        \label{fig:difference_norm}
    \end{subfigure}
    
    \caption{\textbf{Analysis of a linear transformation matrix $T$.} (a) The matrix closely resembles an identity matrix, with strong diagonal values and minimal off-diagonal values. (b) The accumulated gradient is concentrated along the diagonal, emphasizing the importance of training these elements for effective adaptation. 
    }
    \label{fig:observation2-2}
    	\vspace{-0.3cm}   
\end{figure}

In summary, our analysis reveals that Projectors---rather than the SSM---are crucial for learning downstream tasks in Mamba architecture. 
However, Projectors in Mamba represent a large fraction of total parameters, accounting for approximately 65.33\% of the model’s parameters. Although direct fine-tuning of Projectors is effective, training such a large number of parameters is challenging. This observation motivates the need for PEFT methods that indirectly optimizes pretrained Projectors, preserving their capabilities while requiring fewer learnable parameters.

\begin{figure*}[t]
\centering 
    \includegraphics[width=\linewidth,trim={0cm 0cm 0cm 0cm},clip]{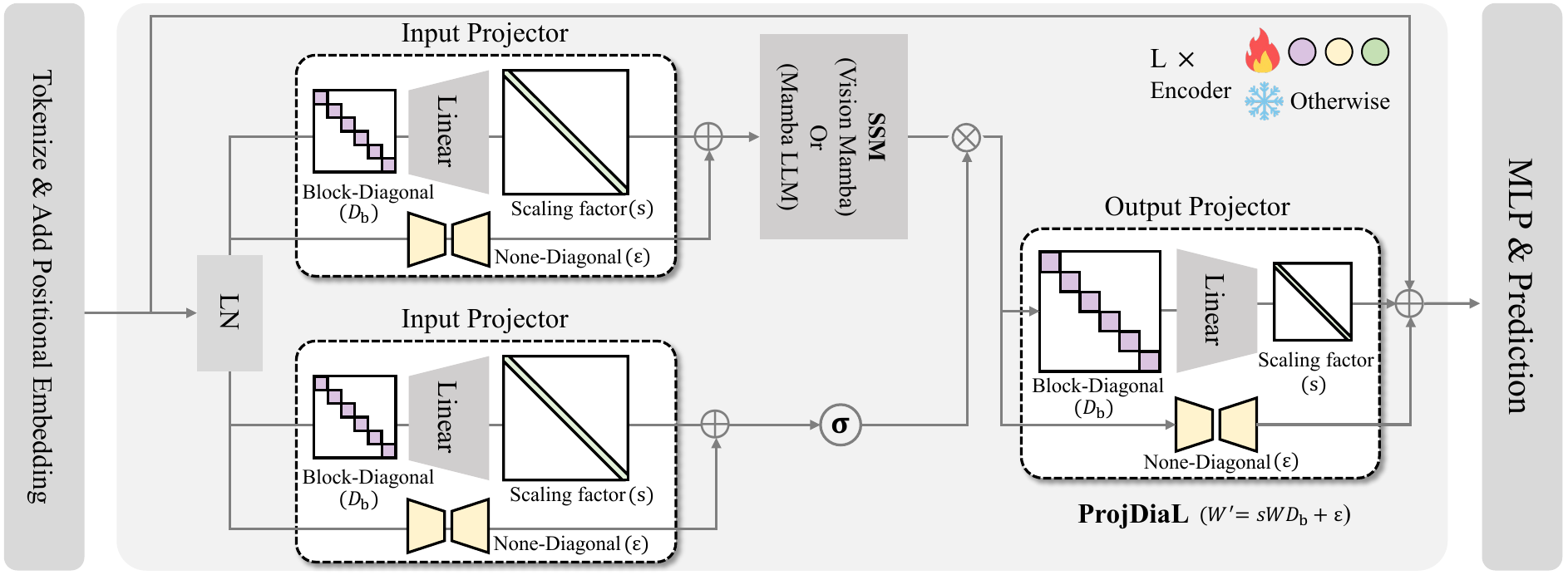} 
	\vspace{-0.5cm}   
	\caption{
\textbf{Overview of ProDiaL Architecture for Efficient Parameter Tuning in Mamba Models}:
A detailed structure of ProDiaL's approach to fine-tuning Mamba architecture by focusing on Projector transformations. ProDiaL selectively updates the diagonal($D_b$) and non-diagonal($\epsilon$) matrices in Projectors, enabling efficient learning with minimal parameters.} 
	\label{fig:teasor}
    	\vspace{-0.3cm}   
\end{figure*}

\subsection{ProDiaL: Projector-targeted Diagonal-centric Linear Transformation}

\subsubsection{Primarily Training Diagonal Entries in $T$.}
Building on our observation that Projectors are essential for transfer learning in Mamba architecture, we analyze the relationship between the pretrained Projector weight $W \in \mathbb{R}^{d_{out} \times d_{in}}$ and fine-tuned Projector weight $W' \in \mathbb{R}^{d_{out} \times d_{in}}$ to gain deeper insights. Firstly, rather than respresenting a relationship between $W$ and $W'$ as $W'=W+\Delta W$ following~\cite{hu2021lora,liu2024dora,kopiczko2023vera}, we approach this from an uncommon perspective by interpreting the relationship as linear and formulate it as follows:
\vspace{-2mm}
\begin{equation}
\vspace{-1mm}
    W'=WT,
\end{equation}
where $T \in \mathbb{R}^{d_{in} \times d_{in}}$ represents a linear transformation matrix representing the relationship between the fine-tuned and pretrained Projector weights. Given the fine-tuned and pretrained Projector weights, we can deterministically calculate $T_{det}$ using the pseudo-inverse of $W$ as follows:
\begin{equation}
    T_{det}=W^{-1}W',
\end{equation}
where $W^{-1}$ represents the pseudo-inverse of $W$.

As shown in ~\cref{fig:diagonal_analysis}, the matrix $T_{det}$ closely resembles an identity matrix, with high values (close to 1) along the diagonal and near-zero values elsewhere. This pattern suggests that the fine-tuned Projector weight can be approximated as a linear transformation of the pretrained Projector weight, primarily highlighting diagonal entries. This structure is consistently observed across both Projectors, all layers, and both Mamba LLM and Vision models.

In addition, we measure the $L_1$ norm of the difference between the identity matrix $\mathbb{I}$ and fully fine-tuned matrix $T$ (with $W$ frozen) to evaluate the relative importance of diagonal versus off-diagonal entries by the extent of accumulated gradients during training. As visualized in ~\cref{fig:difference_norm}, the diagonal entries exhibit large $L_1$ norm differences from $\mathbb{I}$, while the non-diagonal entries show only minor differences. This finding indicates that the gradients of diagonal entries are accumulated during training, and then primarily training the diagonal entries is critical to making $T$ an ideal transformation matrix that approximates $W$ to $W'$.

These analyses demonstrate that indirectly updating the pretrained Projector weight $W$ can be achieved through linear transformation, with focusing on training the diagonal entries to effectively approximate $W$ to $W'$. This provides a key insight for our novel PEFT method targeting Projectors in the Mamba architecture. Further analysis of the diagonal entries in $T$ is provided in \cref{sec:analysis}.

However, as indicated by the ablation study results in~\cref{table:Vim ablation} and the presence of some large values at the off-diagonal entries shown in~\cref{fig:difference_norm}, training off-diagonal entries together with diagonal entries further enhance performance. 
Consequently, our proposed method incorporates both entries in $T$ but trains them separately, reflecting this observation in our approach.

\subsubsection{Decomposing Diagonal and Off-diagonal Entries.}
Despite our observations highlight the importance of diagonal entries in linear transformation matrix, applying the existing PEFT methods like LoRA~\cite{hu2021lora} and DoRA~\cite{liu2024dora} to Projectors cannot fully consider these insights. Therefore, we propose a novel Projector-targeted PEFT method, which is specialized for Mamba architecture, called \textbf{Pro}jector-targeted \textbf{Dia}gonal-centric \textbf{L}inear Transformation (\textbf{ProDiaL}). The overall framework of ProDiaL is illustrated in ~\cref{fig:teasor}.

Considering the different scales of the diagonal and non-diagonal components, as shown in ~\cref{fig:difference_norm}, we decompose the linear transformation matrix $T$ into a diagonal matrix $D \in \mathbb{R}^{d_{in} \times d_{in}}$ and a off-diagonal matrix $b \in \mathbb{R}^{d_{in} \times d_{in}}$. Given that the values of $b$ are small, we simplify the term $Wb$ as $\epsilon$. This decomposition can be expressed as follows: 

\begin{equation}
\begin{split}
W'=WT &=W(D+b)\\ 
&=WD+Wb\\
&=WD+\epsilon.  
\end{split}
\label{eq:decomposition}
\end{equation}

This decomposition forms main strategy of ProDiaL: we freeze the pretrained Projector weight $W$ and fine-tune only the diagonal matrix $D$ and off-diagonal matrix $\epsilon$, indirectly updating $W$ towards the fine-tuned Projector weight $W'$.

Specifically, to enhance flexibility, we replace the diagonal matrix $D$ with a block-diagonal matrix $D_b$ and introduce a learnable scaling parameter $s$, as inspired by~\cref{fig:training_analysis}. This allows subtle transformations, such as minor rotations, beyond simple scaling of Projector weights. 
For the off-diagonal matrix $\epsilon$, which has same dimensions as $W$ and $W'$ but is relatively less important than the diagonal entries, we apply LoRA with a low-rank structure to achieve parameter efficiency.
In~\cref{fig:teasor}, the colored parts indicate the learnable parameters in ProDiaL, where only $D_b$, $\epsilon$, and $s$ are trained while $W$ remains unchanged.

In addition, inspired by \cref{fig:difference_norm}, we indirectly optimize our block-diagonal matrix $D_b$ by learning the difference between an identity matrix $\mathbb{I}$ and an auxiliary block-diagonal matrix $D_a \in \mathbb{R}^{d_{in} \times d_{in}}$. 
Consequently, our ProDiaL method is formulated as follows:

\begin{equation}
    \begin{split}
        & W'=sWD_b +\epsilon, \\
    & D_b =[\mathbb{I}-relu(\mathbb{I}*D_a)] + (\textbf{1}-\mathbb{I})*D_a,  \\
    & D_a = diag(x_1, x_2, ..., x_n), \\
    & \epsilon = B_\epsilon A_\epsilon, 
    \end{split}
\end{equation}
where $s \in \mathbb{R}^{d_{out}}$ is an learnable scaling parameter, ${\{x\}}^{n}_{i=1} \in \mathbb{R}^{(d_{in}/r_b) \times (d_{in}/r_b)}$ are small matrices forming $D_a$, and 
$A_\epsilon \in \mathbb{R}^{r_\epsilon \times d_{in}}$ and $B_\epsilon \in \mathbb{R}^{d_{out} \times r_\epsilon}$ are low-rank matrices employed through LoRA to train $\epsilon$. The detailed algorithm of our ProDiaL including explanation of $r_b$ is provided in Section A of the Supplementary Material.

With these approaches, ProDiaL provides flexibility in controlling the number of learnable parameters by adjusting the block size $r_b$ in $D_a$ and the low rank value $r_\epsilon$ of $\epsilon$. The experiment with varying values of $r_b$ and $r_\epsilon$ are reported in Section B of the Supplementary Material. Furthermore, as \cref{table:main_results} demonstrated, ProDiaL is effective regardless of whether it targets only the Input Projector or Output Projector, achieving comparable performance to fine-tuning both Projectors. After the training, ProDiaL stores only the updated $W$, transformed with matrices $D_b$ and $\epsilon$. Consequently, there is no need to store additional parameters used during training, allowing ProDiaL to follow the advantage of LoRA.

\begin{table*}[t]
    \centering
    \setlength{\tabcolsep}{3pt}
    \scalebox{0.75}{
    \begin{tabular}{y{20}y{70}x{65}x{65}x{65}x{65}x{35}x{60}x{60}x{60}x{35}}
    \arrayrulecolor{black} \toprule
    & & \multicolumn{5}{c}{\textbf{Mamba LLM}} & \multicolumn{4}{c}{\textbf{Vision Mamba}} \\
    \arrayrulecolor{black} \cmidrule(lr){3-7} \cmidrule(lr){8-11}
    & Method & HellaSwag & Winogrande & ARC-E & ARC-C & Avg &  StanfordCars & Caltech & Flowers & Avg. \\ \arrayrulecolor{black} \midrule
    \multirow{5.5}{*}{\rotatebox{90}{\textbf{Baselines}}} 
    & \textcolor{gray}{Full-FT} & \textcolor{gray}{38.23 \footnotesize{(130.00M)}} & \textcolor{gray}{53.12 \footnotesize{(130.00M)}} & \textcolor{gray}{53.54 \footnotesize{(130.00M)}} & \textcolor{gray}{28.84 \footnotesize{(130.00M)}} & \textcolor{gray}{43.43} & \textcolor{gray}{90.06 \footnotesize{(7.00M)}} & \textcolor{gray}{92.86 \footnotesize{(7.00M)}} & \textcolor{gray}{92.05 \footnotesize{(7.00M)}} & \textcolor{gray}{91.66} \\
    \arrayrulecolor{lightgray} \cmidrule(lr){2-11}
    & Linear Probing & - & - & - & - & - & 57.46 \footnotesize{(0.04M)} & 91.10 \footnotesize{(0.02M)} & 59.90 \footnotesize{(0.02M)} & 69.49 \\
    \arrayrulecolor{lightgray} \cmidrule(lr){2-11}
    & BitFit \cite{zaken2021bitfit} & 35.69 \footnotesize{(0.07M)} & 53.12 \footnotesize{(0.07M)} & 52.86 \footnotesize{(0.07M)} & 26.88 \footnotesize{(0.07M)} & 42.14 & 65.51 \footnotesize{(0.08M)} & 93.71 \footnotesize{(0.06M)} & 78.84 \footnotesize{(0.06M)} & 79.35\\     \arrayrulecolor{lightgray} \cmidrule(lr){2-11}
    & Strong \cite{halloran2024mamba} & 38.66 \footnotesize{(3.80M)} & 53.04 \footnotesize{(3.80M)} & 54.17 \footnotesize{(3.80M)} & 28.67 \footnotesize{(3.80M)} & 43.64 & 84.78 \footnotesize{(0.96M)} & 95.70 \footnotesize{(0.94M)} & 86.76 \footnotesize{(0.94M)} & 89.08 \\     \arrayrulecolor{black} \midrule
    \multirow{5.5}{*}{\rotatebox{90}{\textbf{Both-Proj}}} 
    & \textcolor{gray}{FT}  & \textcolor{gray}{40.18 \footnotesize{(84.94M)}} & \textcolor{gray}{52.57 \footnotesize{(84.94M)}} & \textcolor{gray}{54.38 \footnotesize{(84.94M)}} & \textcolor{gray}{29.52 \footnotesize{(84.94M)}} & \textcolor{gray}{44.16} & \textcolor{gray}{89.67 \footnotesize{(5.35M)}} & \textcolor{gray}{95.01 \footnotesize{(5.33M)}} & \textcolor{gray}{92.00 \footnotesize{(5.33M)}} & \textcolor{gray}{92.22} \\
    \arrayrulecolor{lightgray} \cmidrule(lr){2-11}
    & LoRA & 38.33 \footnotesize{(2.36M)} & 53.12 \footnotesize{(2.36M)} & 53.87 \footnotesize{(2.36M)} & \textbf{29.52} \footnotesize{(2.36M)} & 43.71 & 85.06 \footnotesize{(0.63M)} & 96.01 \footnotesize{(0.61M)} & 87.32 \footnotesize{(0.61M)} & 89.46 \\    \arrayrulecolor{lightgray} \cmidrule(lr){2-11}
    & DoRA & 38.13 \footnotesize{(2.45M)} & 52.88 \footnotesize{(2.45M)} & 54.12 \footnotesize{(2.45M)} & 28.75 \footnotesize{(2.45M)} & 43.47 & 85.18 \footnotesize{(0.69M)} & 96.09 \footnotesize{(0.65M)} & 86.60 \footnotesize{(0.65M)} & 89.29 \\
    \arrayrulecolor{lightgray} \cmidrule(lr){2-11}
    & {ProDiaL} & \textbf{38.92} \footnotesize{(2.42M)} &  \textbf{53.28} \footnotesize{(2.42M)} & \textbf{55.18} \footnotesize{(2.38M)} &   28.84 \footnotesize{(2.38M)} & {\textbf{44.06}} &  \textbf{85.38} \footnotesize{(0.67M)} &  \textbf{96.24} \footnotesize{(0.65M)} & \textbf{88.00} \footnotesize{(0.65M)} & {\textbf{89.87}} \\    \arrayrulecolor{gray} \midrule
    \multirow{5.5}{*}{\rotatebox{90}{\textbf{In-Proj}}} 
    & {\textcolor{gray}{FT}}  & \textcolor{gray}{40.36 \footnotesize{(56.62M)}} & \textcolor{gray}{53.20 \footnotesize{(56.62M)}} & \textcolor{gray}{54.59 \footnotesize{(56.62M)}} & \textcolor{gray}{29.61 \footnotesize{(56.62M)}} & {\textcolor{gray}{44.44}} & \textcolor{gray}{89.62 \footnotesize{(3.58M)}} & \textcolor{gray}{95.24 \footnotesize{(3.56M)}} & \textcolor{gray}{91.02 \footnotesize{(3.56M)}} & {\textcolor{gray}{91.96}} \\     \arrayrulecolor{lightgray} \cmidrule(lr){2-11}
    & {LoRA}  & \textbf{38.46} \footnotesize{(1.48M)} & 52.80 \footnotesize{(1.48M)} & 53.87 \footnotesize{(1.48M)} & 28.41 \footnotesize{(1.48M)} & {43.39} & 82.12 \footnotesize{(0.41M)} & 95.78 \footnotesize{(0.39M)} & 85.71 \footnotesize{(0.39M)} & {87.87}\\    \arrayrulecolor{lightgray} \cmidrule(lr){2-11}
    & {DoRA} & 38.08 \footnotesize{(1.55M)} & 52.64 \footnotesize{(1.55M)} & 54.04 \footnotesize{(1.55M)} & 28.50 \footnotesize{(1.55M)} & {43.32} & 82.17 \footnotesize{(0.43M)} & 95.55 \footnotesize{(0.41M)} & 85.95 \footnotesize{(0.41M)} & {87.89} \\    \arrayrulecolor{lightgray} \cmidrule(lr){2-11}
    & {ProDiaL} & 38.41 \footnotesize{(1.49M)} &   \textbf{52.96} \footnotesize{(1.49M)} &   \textbf{54.50} \footnotesize{(1.25M)} &   \textbf{29.61} \footnotesize{(1.25M)} &   {\textbf{43.87}} &   \textbf{82.45} \footnotesize{(0.42M)} &   \textbf{95.93} \footnotesize{(0.41M)} &   \textbf{85.97} \footnotesize{(0.33M)} & {\textbf{88.12}} \\
    \arrayrulecolor{gray} \midrule
    \multirow{5.5}{*}{\rotatebox{90}{\textbf{Out-Proj}}} 
    & {\textcolor{gray}{FT}} & \textcolor{gray}{40.89 \footnotesize{(28.31M)}} & \textcolor{gray}{52.80 \footnotesize{(28.31M)}} & \textcolor{gray}{55.09 \footnotesize{(28.31M)}} & \textcolor{gray}{29.27 \footnotesize{(28.31M)}} & {\textcolor{gray}{44.51}} & \textcolor{gray}{88.86 \footnotesize{(1.81M)}} & \textcolor{gray}{95.63 \footnotesize{(1.77M)}} & \textcolor{gray}{91.45 \footnotesize{(1.77M)}} & {\textcolor{gray}{91.98}} \\    \arrayrulecolor{lightgray} \cmidrule(lr){2-11}
    & {LoRA}  & 37.30 \footnotesize{(0.89M)} & 53.12 \footnotesize{(0.89M)} & 53.66 \footnotesize{(0.89M)} & 28.54 \footnotesize{(0.89M)} & {43.08} & 77.81 \footnotesize{(0.26M)} & 95.40 \footnotesize{(0.24M)} & 80.60 \footnotesize{(0.24M)} & {84.60} \\    \arrayrulecolor{lightgray} \cmidrule(lr){2-11}
    & {DoRA} & 37.19 \footnotesize{(0.90M)} & 52.88 \footnotesize{(0.90M)} & 53.66 \footnotesize{(0.90M)} & 28.67 \footnotesize{(0.90M)} & {43.10} & 77.70 \footnotesize{(0.28M)} & 95.47 \footnotesize{(0.26M)} & 80.97 \footnotesize{(0.26M)} & {84.71} \\    \arrayrulecolor{lightgray} \cmidrule(lr){2-11}
    & {ProDiaL} &   \textbf{38.19} \footnotesize{(0.92M)} &   \textbf{53.75} \footnotesize{(0.92M)} &   \textbf{54.84} \footnotesize{(0.90M)} &   \textbf{30.80} \footnotesize{(0.90M)} &   {\textbf{44.40}} &   \textbf{78.00} \footnotesize{(0.27M)} &   \textbf{95.55} \footnotesize{(0.25M)} &   \textbf{81.90} \footnotesize{(0.25M)} &   {\textbf{85.15}} \\ 
    \arrayrulecolor{black} \bottomrule
    \end{tabular}
    }
    \caption{\textbf{Comparison of accuracy(\%) as a downstream task performance in Mamba LLM and Vision Mamba across various datasets.} ProDiaL methods (in Both-Proj, In-Proj, and Out-Proj) consistently demonstrate superior accuracy across both language and vision tasks, achieving similar parameter efficiency compared to other methods.}
    \label{table:main_results}
\end{table*}

\section{Experiment}
In this section, we evaluate the effectiveness of applying PEFT methods to Projectors within the Mamba architecture using our ProDiaL across various downstream tasks.

\subsection{Experiment Setup}

To evaluate the effectiveness of fine-tuning projectors using our ProDiaL, we conducted experiments on Mamba-130M~\cite{gu2023mamba}, which is based on the Mamba 1 architecture for Mamba LLM, and on Vim-tiny~\cite{zhu2024vision}, which is built upon the Mamba 2 architecture~\cite{dao2024transformers} for Vision Mamba.
Specifically, we optimize Mamba LLM pre-trained on the PILE dataset~\cite{gao2020pile} to reasoning tasks~\cite{zellers2019hellaswag,sakaguchi2021winogrande,clark2018think} and Vision Mamba model pre-trained on the ImageNet dataset~\cite{deng2009imagenet} to other classification tasks~\cite{krause20133d,li_andreeto_ranzato_perona_2022,nilsback2008automated}. For Mamba LLM, we measure accuracy by sampling $N$ checkpoints of fine-tuned model weights every $M$ iterations, selecting the checkpoint with the highest test accuracy among the samples. The values of $N$ and $M$ vary across datasets. For Vision Mamba, we measure accuracy of the last checkpoint on the test set. For comparison, we established baselines using Full Fine-Tuning (Full-FT) and existing PEFT methods targeting different parameters, such as BitFit~\cite{zaken2021bitfit} and Strong~\cite{halloran2024mamba}. 
All experiments details, including the details for analysis experiments, are provided in Section D of the Supplementary Material.

\subsection{Experiment Results}

As the first study to report downstream task performance of Mamba LLM and Vision Mamba within Mamba architecture research, we provide extensive experimental results across various datasets, as shown in~\cref{table:main_results}.
Given our experiments yield consistent results across both Mamba LLM and Vision Mamba, we explain the findings comprehensively.

Our results reveal that fine-tuning only the Projectors---or even a single Projector (either input or output)---achieves competitive performance to full fine-tuning, but with a significantly fewer learnable parameters. Even, fully fine-tuning the entire Mamba model or Projectors often leads to overfitting due to the large number of parameters relative to the small size of downstream task data, highlighting the necessity of PEFT methods.

Among existing PEFT methods, LoRA~\cite{hu2021lora} and DoRA~\cite{liu2024dora} applied to Mamba’s Projectors shows impressive downstream performance, demonstrating that high accuracy can be achieved with fewer parameters by targeting Projectors in Mamba. 
For example, compared to Strong~\cite{halloran2024mamba}---which uses LoRA to train $W_x$, both Projectors, and Embeddings---applying LoRA solely to Projectors achieves significant downstream task performance while utilizing only $63.6\%$ of the parameter counts used by Strong for Mamba LLM and $69.8\%$ for Vision Mamba.

Beyond the existing PEFT methods, our ProDiaL, specifically designed based on the analysis of Projectors in Mamba architecture, consistently achieves superior downstream performance with few learnable parameters. 
This results stem not only from targeting Projectors in Mamba architecture but also from highlighting the importance of diagonal entries observed in the linear transformation relationship between pretrained and fine-tuned Projectors.
In ~\cref{table:main_results}, we align the number of parameters of ProDiaL with those of LoRA and DoRA for comparison. More detailed settings, including hyperparameters for $r_b$ and $r_\epsilon$, are provided in Section D of the Supplementary Material.
Furthermore, additional experiments on the Mamba LLM based on the Mamba 2 architecture, along with comprehensive experiment details, are also included in Section C of the Supplementary Material.

\subsection{Results Across Various Model Sizes}

~\Cref{table:LLM_model_size} highlights performance across different model sizes, including Mamba-370M and Mamba-1.4B for Mamba LLM and Vim-small for Vision Mamba, evaluated on the Winogrande~\cite{sakaguchi2021winogrande} and Caltech~\cite{li_andreeto_ranzato_perona_2022} datasets, respectively. 
Consistent with findings from smaller models, fine-tuning only the Projectors—--either individually or jointly—--demonstrates notable effectiveness in adapting to downstream tasks. 
Our ProDiaL, tailored to the Mamba architecture, consistently outperforms other PEFT methods, achieving high accuracy with minimal parameters.
These results emphasize the adaptability and efficiency of our ProDiaL method across model scales, suggesting its potential applicability to other Mamba-based architectures.

\begin{table}[t]
\centering
\scriptsize
\begin{tabular}{c|l|c|c|c}
\hline \hline
\rotatebox{90}{\textbf{}} & \multicolumn{1}{l}{Method} & \multicolumn{1}{c}{Mamba-370M} & \multicolumn{1}{c}{Mamba-1.4B} & \multicolumn{1}{c}{Vim-small} \\ \hline
\multirow{3}{*}{\rotatebox{90}{Base}} 
& Full-FT        & \textcolor{gray}{56.99 (370M)}  & \textcolor{gray}{61.17 (1.40B)} & \textcolor{gray}{94.09 (25.45M)}   \\
& BitFit~\cite{zaken2021bitfit}          & 56.99 (0.20M) & 61.25 (0.39M) & 96.62 (0.11M)  \\
& Strong~\cite{halloran2024mamba}  & 57.14 (8.19M) & 61.80 (0.06B) & 96.47 (1.85M)  \\ \hline

\multirow{4}{*}{\rotatebox{90}{Both-Proj}}
& FT         & \textcolor{gray}{57.22 (302M)} & \textcolor{gray}{61.72 (1.21B)} & \textcolor{gray}{94.94 (21.27M)}  \\
& LoRA       & 56.75 (6.29M) & 61.72 (0.05B) & 96.85 (1.22M)  \\
& DoRA       & 56.99 (6.54M) & 61.72 (0.05B) & 96.85 (1.27M)  \\
& ProDiaL       & \textbf{57.06} (5.75M) & \textbf{61.96} (0.05B) & \textbf{97.16} (1.32M)   \\ \hline

\multirow{4}{*}{\rotatebox{90}{In-Proj}} 
& FT     & \textcolor{gray}{57.06 (201M)} & \textcolor{gray}{61.56 (0.81B)} & \textcolor{gray}{95.17 (14.20M)}  \\
& LoRA   & 56.75 (3.93M) & \textbf{61.56} (0.03B) & \textbf{97.16} (0.88M)  \\
& DoRA   & \textbf{57.22} (4.13M) & 61.48 (0.03B) & \textbf{97.16} (0.91M)  \\
& ProDiaL   & \textbf{57.22} (3.74M) & \textbf{61.56} (0.03B) & 97.09 (0.82M)  \\  \hline

\multirow{4}{*}{\rotatebox{90}{Out-Proj}} 
& FT   & \textcolor{gray}{57.22 (101M)} & \textcolor{gray}{61.72 (0.40B)}  & \textcolor{gray}{95.86 (7.12M)}   \\
& LoRA & 56.91 (2.36M) & 61.56 (0.02B) & 96.70 (0.48M)     \\
& DoRA  & 57.22 (2.41M) & 61.48 (0.02B) & 96.78 (0.59M)  \\
& ProDiaL & \textbf{57.30} (2.02M) & \textbf{61.80} (0.02B)  & \textbf{96.85} (0.51M)  \\  \hline \hline
\end{tabular}
\vspace{-3mm}
\caption{\textbf{Comparison of downstream task performance (\%) across various model sizes with different fine-tuning methods.}}
\label{table:LLM_model_size}
\end{table}

\begin{figure}[!t]
    \centering
    \includegraphics[width=0.9\linewidth]{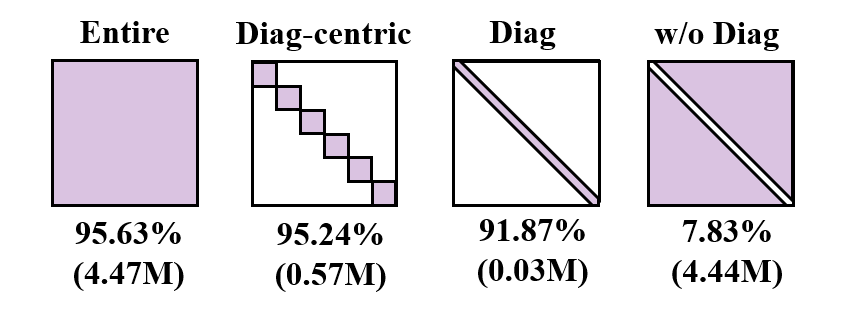}
    \vspace{-3mm}
    \caption{\textbf{Performance comparison across four $T$ configurations in Mamba model.} The diagonal-centric approach achieves near-optimal performance with a significantly reduced parameter number ($0.57M$), supporting the validity of our ProDiaL. 
    }
    \label{fig:training_analysis}
    \vspace{-3mm}
\end{figure}

\subsection{Analysis} \label{sec:analysis}

\noindent\textbf{Q: Training Only Diagonal Components is Sufficient?}
\vspace{1mm}

\noindent To empirically verify that training primarily the diagonal elements of a linear transformation matrix $T$ is sufficient, we conduct a experiment, comparing four approaches: (1) directly fine-tuning $T$, (2) fine-tuning only the diagonal-centric matrix, (3) fine-tuning only the diagonal matrix, (4) fine-tuning only the off-diagonal matrix. The four approaches and their performances are illustrated in~\cref{fig:training_analysis}. 
The results show that fine-tuning only the diagonal matrix achieves significant performance despite of its small number of parameters. Expanding the diagonal matrix to a diagonal-centric matrix further enhances performance, achieving results comparable to directly fine-tuning $T$. In contrast, fine-tuning only the off-diagonal matrix significantly degrades performance.
Consequently, this analysis emphasizes the importance of primarily training the diagonal elements in the linear transformation matrix $T$, supporting our strategy in ProDiaL.\\

\vspace{1mm}

\noindent\textbf{Q: Diagonal is Important Only in Projectors?}
\vspace{1mm}

\noindent Our observations highlight that fine-tuning only the diagonal entries in the transformation matrix \( T \) effectively approximates pretrained projector weights \( W \) to fine-tuned weights \( W' \). However, it is unclear whether the importance of the diagonal entries is unique to Projectors or extends to all linear layers in the Mamba architecture.

To investigate this, we insert learnable diagonal matrices to the SSM’s linear layers (\( W_x \) and \( W_\Delta \)) and evaluate their impact. ~\Cref{table: Projector analysis2} shows that in the SSM, fine-tuning only the diagonal matrix degrades performance due to the SSM's already small parameter count, limiting effective learning. In contrast, for Projectors, diagonal-only fine-tuning improves performance by avoiding overfitting, a common issue with large parameter numbers and limited data. Furthermore, applying diagonal matrices to both SSM and Projectors confirms that indirectly fine-tuning with a diagonal matrix works best for components with a large number of parameters, as it reduces overfitting.

Consequently, diagonal fine-tuning benefits any linear layer with large number of parameters, but in Mamba architecture, Projectors are the solely component with large number of parameters, making them the ideal target for this parameter-efficient approach.\\

\begin{table}[!t] 
\centering
\begin{tabular}{c|c}
\hline
 Target   & Accuracy (\%) \\ \hline
 SSM (FT) &  94.01 \\ 
 SSM (Diag) & 93.55 \\ \hline
 Proj (FT) &  95.01 \\ 
 Proj (Diag) & 95.24 \\ \hline
 SSM+Proj (FT) & 94.55 \\ 
 SSM+Proj (Diag) & 95.55 \\ \hline
 \end{tabular}
 \vspace{-2mm}
\caption{\textbf{Analysis of Diagonal-only Fine-tuning.}  Projectors benefit more than SSM from diagonal-only fine-tuning.}
\label{table: Projector analysis2}
\end{table}

\begin{table}[!t]
\centering
\small
\begin{tabular}{lcc}
\hline
Target Components & \# of parameters & Accuracy \\ \hline
Full-FT                         & 85.9M   & 92.83  \\
Attention                       & 28.4M   & 93.07 \\
FFN                             & 56.8M  & 92.89  \\  
Attention (LoRA)                & 0.67M   & 92.60 \\
FFN (LoRA)                      & 0.81M  & 92.51  \\  
Attention+FFN (LoRA)            & 1.40M  & 92.91  \\ \hline
\end{tabular}
\vspace{-2mm}
\caption{\textbf{Performance(\%) of Transformer Components under Fine-Tuning.} Attention module achieves slightly higher performance than FFN, but yields similar performance levels.} 
\label{table:observation Transformers}
\end{table}

\begin{table}[!t] 
\centering
\begin{tabular}{l|c}
\hline
 Method & Accuracy (\%)  \\ \hline
 Proj-$D_b$ & 95.24 (0.57M) \\
 Proj-$D_b$+$\epsilon$ & 96.16 (0.62M) \\ 
 Proj-$D_b$+$\epsilon$+$s$ (ProDiaL) & 96.24 (0.65M) \\ \hline
 \end{tabular}
 \vspace{-2mm}
\caption{\textbf{Ablation Study on ProDiaL Components.}}
\label{table:Vim ablation}
 \vspace{-2mm}
\end{table}

\noindent\textbf{Q: How Does Mamba PEFT Differ from Transformers?}

\noindent To further investigate whether non-attention modules in Transformers can effectively capture downstream task knowledge as Projectors do in Mamba, we conduct an experiment by applying LoRA~\cite{hu2021lora} to each component in the ViT model~\cite{dosovitskiy2020image}. 
As shown in~\cref{table:observation Transformers}, both attention modules and the Feed Forward Network (FFN)---a non-attention module---achieve comparable performance, with a slight advantage for the attention modules. This result aligns with Mamba, where fine-tuning non-SSM components (Projectors) is effective for downstream tasks. 

However, in Transformers, jointly fine-tuning both attention and FFN together synergistically enhance the downstream task performance. In contrast, in Mamba architecture, fine-tuning both SSM and Projectors together degrades performance, as shown in~\cref{table:observation vim,table:observation LLM}. In other words, Mamba benefits from fine-tuning only Projectors without the SSM, whereas Transformers achieve optimal performance by fine-tuning both Attention and FFN. 
This suggests that fine-tuning only Projectors for downstream tasks is uniquely beneficial to Mamba architecture, distinguishing the characteristics of Mamba from Transformers.

\subsection{Ablation Study}

Our ProDiaL consists of a learnable block-diagonal matrix $D_b$, a non-diagonal matrix $\epsilon$, and a scaling factor $s$, which together enable effective learning of downstream tasks in Mamba architecture. To analyze the contribution of each component, we perform an ablation study by adding components incrementally. As shown in \cref{table:Vim ablation}, fine-tuning only the diagonal-centric matrices achieves significant downstream task performance. While fine-tuning only diagonal-centric matrices proves effective, fine-tuning both the non-diagonal matrices and the diagonal matrices yields even higher downstream task performance. In addition, inspired by \cite{qiu2023controlling}, incorporating scaling factors at the output stage further enhances downstream task performance. 
These results demonstrate that each component of ProDiaL contributes meaningfully to maximizing downstream task performance within the Mamba architecture.

\section{Limitations}
Our ProDiaL primiarily relies on LoRA~\cite{hu2021lora} to efficiently train off-diagonal matrix $\epsilon$. Developing a method specifically for optimizing $\epsilon$ could further enhance performance, and is an interesting direction for future work.

\section{Conclusion}

In this work, we reveal that Projectors play a critical role in transfer learning within Mamba architecture. Based on our observation that fine-tuned Projectors can be approximated through a near-diagonal linear transformation, we propose ProDiaL, a novel PEFT method specifically designed for Mamba architecture. Rather than fine-tuning SSM, ProDiaL targets Projectors, optimizing them indirectly through a diagonal-centric linear transformation. This enables ProDiaL to achieve superior performance while fine-tuning less than 1\% of the model parameters. Our experiments across both vision and language Mamba models demonstrate ProDiaL’s effectiveness, establishing a new benchmark for PEFT in Mamba-based architecture.

{
    \small
    \bibliographystyle{ieeenat_fullname}
    \bibliography{main}
}

\clearpage
\appendix
\setcounter{page}{1}
\setcounter{section}{0}
\renewcommand{\thefigure}{S\arabic{figure}}
\renewcommand{\thetable}{S\arabic{table}}
\setcounter{figure}{0}
\setcounter{table}{0}
\maketitlesupplementary

\section{Algorithms of ProDiaL}

\begin{algorithm}[!t]
\caption{ProDiaL: Projector-targeted Diagonal-centric Linear Transformation}\label{alg:prodial}
\begin{algorithmic}[1]
\Require Pretrained Projector weight $W$, Downstream task data $X$, Learning rate $\eta$, Hyperparameter $r_b, r_\epsilon$ for ProDiaL.
\Ensure Fine-tuned Projector weight $W'$
\State Set the hyperparameter $r_b, r_\epsilon$ and learning rates $\eta$
\State Initialize $r_b$ small block matrices $x_1, ..., x_{r_b} \in \mathbb{R}^{(d_{in}/r_b) \times (d_{in}/r_b)}$ with identity matrix
\State Construct $D_a = diag(x_1, ..., x_{r_b})$
\State Construct $D_b = [\mathbb{I}-relu(\mathbb{I}*D_a)]+(\textbf{1}-\mathbb{I})*D_a$
\State Initialize scaling factor $s$ with one vector
\State Initialize low-rank matrices $A_\epsilon$ with random noise $\mathcal{N}(0,\sigma^2)$ and $B_\epsilon$ with zeros
\While{not converged}
    \State Sample a mini-batch $X_\text{batch}$ from the downstream task data
    \State Compute the intermediate transformation: $W' = sWD_b + B_\epsilon A_\epsilon$
    \State Perform forward pass with $W'$ on $X_\text{batch}$
    \State Compute loss $\mathcal{L}(W')$ based on task objective
    \State Backpropagate gradients $\nabla_{D_b} \mathcal{L}, \nabla_{A_\epsilon} \mathcal{L}, \nabla_{B_\epsilon} \mathcal{L}$
    \State Update $D_b \gets D_b - \eta \cdot \nabla_{D_b} \mathcal{L}$
    \State Update $A_\epsilon \gets A_\epsilon - \eta \cdot \nabla_{A_\epsilon} \mathcal{L}$
    \State Update $B_\epsilon \gets {B_\epsilon} - \eta \cdot \nabla_{B_\epsilon} \mathcal{L}$
\EndWhile
\State Compute final fine-tuned weight: $W' = sWD_b + B_\epsilon A_\epsilon$
\State \Return $W'$
\end{algorithmic}
\end{algorithm}

The ProDiaL (Projector-targeted Diagonal-centric Linear Transformation) is a parameter-efficient fine-tuning (PEFT) method specifically designed for Mamba architecture’s Projectors. It efficiently and effectively adapts pretrained Projector weights W to downstream tasks via a diagonal-centric linear transformation, significantly minimizing the number of learnable parameters. \Cref{alg:prodial}  presents the full detailed ProDiaL algorithm.

\begin{figure}[!t]
    \centering
    \hspace{1cm}
    \includegraphics[width=0.6\linewidth]{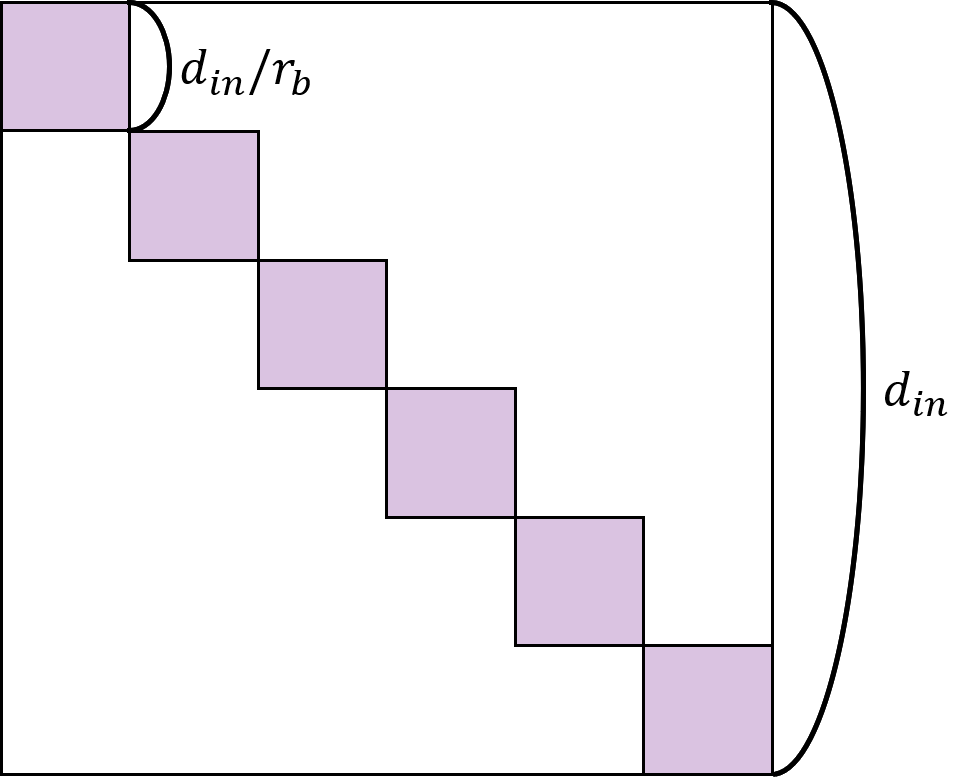}
\caption{\textbf{Block-Diagonal Matrix Design for ProDiaL.} The diagram illustrates the block-diagonal structure of the transformation matrix $D_b$, with $r_b$ controlling the size and number of small block matrices ($x_1, ..., x_{r_b}$). As $r_b$ increases, the block size decreases.}

    \label{fig:block-diagonal}
\end{figure}

The algorithm begins with the initialization of key components and hyperparameters. Firstly, small block matrices $x_1, ..., x_{r_b}$ are initialized as identity matrices and used to construct the auxiliary block-diagonal transformation matrix $D_a$. The hyperparameter $r_b$ controls the size and the number of these small block matrices. As $r_b$ increases, the sizes of each small block decreases, and the number of small blocks increases, as illustrated in~\cref{fig:block-diagonal}. When $r_b$ equals the input dimension $d_{in}$, the small block matrices form a perfectly diagonal matrix. With these small block matrices, the block-diagonal matrix $D_a$ is constructed. To facilitate  faster convergence, ProDiaL trains the difference between identity matrix and diagonal entries. Then, the final diagonal-centric linear transformation matrix $D_b$ is derived from $D_a$. Additionally, a scaling factor $s$ is initialized as a vector of ones, and low-rank matrices $A_\epsilon$ and $B_\epsilon$ are initialized with random noise and zeros, respectively, to address off-diagonal matrices. 

After the initialization and settings, the fine-tuning process iteratively refines these components using mini-batches of downstream task data. For each mini-batch, the algorithm computes the updated Projector weights $W'$ as a combination of the scaled diagonal transformation $sWD_b$ and the low-rank adjustment for off-diagonal matrices $B_\epsilon A_\epsilon$. A forward pass is performed using $W'$, followed by the computation of the task-specific loss $L(W')$. Next, the gradients obtained from $L(W')$ are backpropagated to update $D_b, A_\epsilon, \text{and} B_\epsilon$ via gradient descent. This process is repeated until convergence, ensuring that the learned transformations effectively adapt $W$ to the downstream task.

Upon convergence, the final fine-tuned weights $W'$ are returned as the output of this algorithm. By employing a block-diagonal structure in $D_b$ and low-rank matrices for $A_\epsilon$ and $B_\epsilon$, ProDiaL achieves strong downstream task performance with parameter efficiency and flexibility. This approach is particularly advantageous for large-scale Mamba-based models in both vision and language domains, where full fine-tuning is computationally challenging.

\section{Trade-off between Performance and Number of Parameters by Controlling $r_b$ and $r_\epsilon$}

Our ProDiaL method offers superior flexibility in determining learnable parameters by controlling the size of small block matrices $(x_1, ..., x_{r_b})$ in the block-diagonal matrix $D_b$ using \(r_b\) and the low-rank value for LoRA using \(r_\epsilon\). To examine how performance and the number of parameters vary depending on \(r_b\) and \(r_\epsilon\), we conducted experiments using the Vim-tiny model~\cite{zhu2024vision} on the Caltech~\cite{li_andreeto_ranzato_perona_2022} and Flowers datasets~\cite{nilsback2008automated}. The hyperparameter \(r_{b1}\) controls the small block size of the Input Projectors, while \(r_{b2}\) controls the small block size of the Output Projectors.

~\Cref{tab:varying hyperparameters} shows the performance and the number of parameters for different values of \(r_b\) and \(r_\epsilon\). Firstly, we confirm that varying the small block sizes (the number of parameters) does not degrade performance. In other words, this demonstrates that it is possible to flexibly adjust the number of parameters by tuning \(r_b\) and \(r_\epsilon\), enabling a trade-off between performance and computational cost (parameter usage). Interestingly, even replacing the block-diagonal matrix with a diagonal matrix---represented by the case \((r_{b1}, r_{b2}) = (192, 384)\)---yields comparable performance with the smallest number of parameters among the same \(r_\epsilon\) values.
Secondly, we observe that the optimal number of parameters for the best performance depends on the dataset. For the Caltech dataset, the highest accuracy is achieved with a relatively small number of parameters, whereas for the Flowers dataset, the best accuracy requires the largest number of parameters. This suggests that simpler datasets (those with higher baseline accuracy) can achieve high performance with fewer parameters, while more complex datasets (those with lower baseline accuracy) need a larger parameters for high performance.

\begin{table}[!t]
\begin{tabular}{cccccc}
\hline
 $r_{b1}$ & $r_{b2}$ & $r_\epsilon$ & Params & Caltech (\%) & Flowers (\%) \\ \hline
 192 & 384 & 16 & 0.65M & 96.24 & 86.96 \\
 32 & 32 & 16 & 0.77M & 95.93 & 89.12 \\
 16 & 32 & 16 & 0.80M & 95.86 & 89.22 \\
 8 & 32 & 16 & 0.85M & 96.01 & 89.51 \\
 32 & 16 & 16 & 0.88M & 95.63 & 87.75 \\
 16 & 16 & 16 & 0.91M & 95.63 & 89.80 \\
 8 & 16 & 16 & 0.96M & 96.24 & 89.90 \\
 32 & 8 & 16 & 1.10M & 95.63 & 87.75 \\
 16 & 8 & 16 & 1.13M & 95.47 & 90.10 \\
 8 & 8 & 16 & 1.19M & 95.70 & \textbf{90.29} \\
 192 & 384 & 8 & 0.35M & 95.70 & 88.63 \\
 32 & 32 & 8 & 0.48M & 96.09 & 87.94 \\
 16 & 32 & 8 & 0.50M & 96.01 & 88.92 \\
 8 & 32 & 8 & 0.56M & \textbf{96.62} & 89.12 \\
 32 & 16 & 8 & 0.59M & 96.24 & 88.33 \\
 16 & 16 & 8 & 0.61M & 96.32 & 89.22 \\
 8 & 16 & 8 & 0.67M & 96.09 & 89.61 \\
 32 & 8 & 8 & 0.81M & 95.70 & 89.80 \\
 16 & 8 & 8 & 0.84M & 96.09 & 89.02 \\
 8 & 8 & 8 & 0.89M & 95.93 & 90.10 \\ \hline
\end{tabular}
\caption{\textbf{Performance Across Varying Hyperparameters.} The table demonstrates the impact of varying block sizes ($r_{b1}, r_{b2}$) and low-rank value ($r_\epsilon$) on the number of parameters and performance for the Caltech and Flowers datasets. Smaller parameter counts perform well on simpler datasets (e.g., Caltech), while larger parameter counts yield better performance on more complex datasets (e.g., Flowers).}

\label{tab:varying hyperparameters}
\end{table}

\section{Additional Experiments}

\subsection{Evaluation on additional datasets}

\begin{table}[!t]
\centering
\small
\begin{tabular}{c|l|c|c}
\hline
\rotatebox{90}{\textbf{}} & \multicolumn{1}{l}{Datasets} & \multicolumn{1}{c}{SIQA} & \multicolumn{1}{c}{OBQA}\\ \hline
\multirow{3}{*}{\rotatebox{90}{\scriptsize Both-Proj}}
& LoRA       & 34.14 (2.36M) & 38.00 (1.18M) \\
& DoRA       & 34.34 (2.45M) & 38.20 (1.27M) \\
& ProDiaL    & \textbf{34.80} (2.51M) & \textbf{39.60} (1.18M) \\ \hline
\multirow{3}{*}{\rotatebox{90}{\scriptsize In-Proj}} 
& LoRA   & 34.80 (1.48M) & 36.40 (0.74M)\\
& DoRA   & 34.39 (1.55M) & 36.80 (0.81M) \\
& ProDiaL   & \textbf{35.47} (1.03M) & \textbf{37.80} (0.89M) \\  \hline
\multirow{3}{*}{\rotatebox{90}{\scriptsize Out-Proj}} 
& LoRA & 33.16 (0.89M) & 33.80 (0.44M) \\
& DoRA  & 33.27 (0.90M) & 34.20 (0.46M) \\
& ProDiaL & \textbf{33.93} (0.90M) & \textbf{35.00} (0.46M) \\  \hline
\end{tabular}
\caption{\textbf{Performance (\%) on SIQA and OBQA Datasets.}}
\label{table:additional_dataset}
\end{table}

To further evaluate the effectiveness of our proposed ProDiaL method, we conducted additional experiments on SIQA~\cite{sap2019socialiqa} and OBQA~\cite{OpenBookQA2018} datasets. These datasets were selected to assess the wider reasoning capabilities of ProDiaL across diverse tasks and domains. As shown in \cref{table:additional_dataset}, our ProDiaL method consistently outperforms both LoRA and DoRA, demonstrating superior performance in line with the results observed on other datasets in the main manuscript. This consistent improvement underscores the robustness of ProDiaL and its ability to effectively adapt across various data distributions and task complexities.

\begin{table*}[t]
    \centering
    \setlength{\tabcolsep}{3pt} 
    \begin{tabular}{y{20}y{70}x{65}x{65}x{65}x{65}x{35}}
    \arrayrulecolor{black} \toprule
    & Method & HellaSwag & Winogrande & ARC-E & ARC-C & Avg \\ \arrayrulecolor{black} \midrule
    & \textcolor{gray}{Full-FT} & \textcolor{gray}{38.23 \footnotesize{(130.00M)}} & \textcolor{gray}{53.12 \footnotesize{(130.00M)}} & \textcolor{gray}{53.54 \footnotesize{(130.00M)}} & \textcolor{gray}{28.84 \footnotesize{(130.00M)}} & \textcolor{gray}{43.43} \\
    \arrayrulecolor{black} \midrule
    \multirow{5.5}{*}{\rotatebox{90}{\textbf{Both-Proj}}} 
    & \textcolor{gray}{FT}  & \textcolor{gray}{38.76 \footnotesize{(90.1M)}} & \textcolor{gray}{53.12 \footnotesize{(90.1M)}} & \textcolor{gray}{50.67 \footnotesize{(90.1M)}} & \textcolor{gray}{28.84 \footnotesize{(90.1M)}} & \textcolor{gray}{42.84} \\     \arrayrulecolor{lightgray} \cmidrule(lr){2-7}
    & LoRA & 38.50 \footnotesize{(2.47M)} & 53.35 \footnotesize{(2.47M)} & 52.36 \footnotesize{(2.47M)} & 30.20 \footnotesize{(2.47M)} & 43.60  \\    \arrayrulecolor{lightgray} \cmidrule(lr){2-7}
    & DoRA & 35.24 \footnotesize{(2.57M)} & 52.01 \footnotesize{(2.57M)} & 47.18 \footnotesize{(2.57M)} & 24.15 \footnotesize{(2.57M)} & 39.65 \\     \arrayrulecolor{lightgray} \cmidrule(lr){2-7}
    & {ProDiaL} & \textbf{38.57} \footnotesize{(2.44M)} &  \textbf{53.83} \footnotesize{(2.00M)} & \textbf{53.03} \footnotesize{(2.33M)} & \textbf{30.46} \footnotesize{(2.22M)} & {\textbf{43.97}}  \\        \arrayrulecolor{gray} \midrule
    \multirow{5.5}{*}{\rotatebox{90}{\textbf{In-Proj}}} 
    & {\textcolor{gray}{FT}}  & \textcolor{gray}{39.89 \footnotesize{(61.8M)}} & \textcolor{gray}{53.35 \footnotesize{(61.8M)}} & \textcolor{gray}{51.56 \footnotesize{(61.8M)}} & \textcolor{gray}{27.22 \footnotesize{(61.8M)}} & {\textcolor{gray}{43.01}} \\     
    \arrayrulecolor{lightgray} \cmidrule(lr){2-7}
    & {LoRA}  & 37.32 \footnotesize{(1.58M)} & 53.43 \footnotesize{(1.58M)} & 52.02 \footnotesize{(1.58M)} & 30.03 \footnotesize{(1.58M)} & 43.20 \\    \arrayrulecolor{lightgray} \cmidrule(lr){2-7}
    & {DoRA} & 35.24 \footnotesize{(1.66M)} & 52.01 \footnotesize{(1.66M)} & 47.18 \footnotesize{(1.66M)} & 24.15 \footnotesize{(1.66M)} & 39.65  \\    \arrayrulecolor{lightgray} \cmidrule(lr){2-7}
    & {ProDiaL} & \textbf{37.91} \footnotesize{(0.98M)} & \textbf{53.75} \footnotesize{(0.89M)} & \textbf{53.03} \footnotesize{(0.98M)} &   \textbf{30.29} \footnotesize{(0.93M)} &   {\textbf{43.75}} \\
    \arrayrulecolor{gray} \midrule
    \multirow{5.5}{*}{\rotatebox{90}{\textbf{Out-Proj}}} 
    & {\textcolor{gray}{FT}} & \textcolor{gray}{40.62 \footnotesize{(28.3M)}} & \textcolor{gray}{53.43 \footnotesize{(28.3M)}} & \textcolor{gray}{54.08\footnotesize{(28.3M)}} & \textcolor{gray}{29.10 \footnotesize{(28.3M)}} & {\textcolor{gray}{44.31}} \\    
    \arrayrulecolor{lightgray} \cmidrule(lr){2-7}
    & {LoRA}  & 37.44 \footnotesize{(0.89M)} & 53.28 \footnotesize{(0.89M)} & 52.65 \footnotesize{(0.89M)} & 28.50 \footnotesize{(0.89M)} & 42.97  \\    \arrayrulecolor{lightgray} \cmidrule(lr){2-7}
    & {DoRA} & 37.44 \footnotesize{(0.90M)} & 53.59 \footnotesize{(0.90M)} & 52.69 \footnotesize{(0.90M)} & 28.92 \footnotesize{(0.90M)} & 43.16  \\    \arrayrulecolor{lightgray} \cmidrule(lr){2-7}
    & {ProDiaL} & \textbf{37.86} \footnotesize{(0.90M)} &   \textbf{53.51} \footnotesize{(0.50M)} &   \textbf{53.45} \footnotesize{(0.68M)} &   \textbf{30.29} \footnotesize{(0.90M)} &   \textbf{43.78}  \\ 
    \arrayrulecolor{black} \bottomrule
    \end{tabular}
    \caption{\textbf{Comparison of accuracy(\%) as a downstream task performance of Mamba2 architecture based LLM across various datasets.} Consistent with Mamba1 architecture, ProDiaL methods (in Both-Proj, In-Proj, and Out-Proj) achieve superior accuracy with smaller or similar number of parameter compared to other methods.}
    \label{table:mamba2_results}
\end{table*}

\subsection{Evaluation on Mamba2 architecture}

To further assess the effectiveness of our ProDiaL method within Mamba 2  architecture~\cite{dao2024transformers}, we conducted additional experiments on LLMs based on the Mamba2-130M. Mamba 2 is an advanced model that addresses the limitations of Mamba 1 by incorporating selective state updates, efficient parallelization, and a lightweight attention mechanism, ensuring efficient and strong performance even with long sequences. Unlike Mamba 1, where Mamba 2 employs a Hybrid Attention Mechanism, the performance of ProDiaL is not fully explored due to its structural differences.
As presented in~\cref{table:mamba2_results}, our ProDiaL method consistently achieves superior accuracy while maintaining a comparable or smaller number of parameters than other approaches. These results demonstrate that ProDiaL effectively enhances downstream task performance in the Mamba 2 architecture, highlighting its strong generalization capability. Details of the hyperparameter configurations, including the block size settings for ProDiaL, are available in~\cref{tab:Mamba2-130M hyperparameters}.

\begin{table*}[!t]
\centering
\small
\begin{tabular}{c|l|c|c||c|c}
\hline
 & Datasets & HellaSwag & {ARC-E} & {StanfordCars} & {Flowers}\\ \hline
\multirow{3}{*}{\rotatebox{90}{\scriptsize Both-Proj}} 
& LoRA       & 38.54$_{\pm 0.18}$ (2.36M) & 54.00$_{\pm 0.11}$ (2.36M) & 85.32$_{\pm 0.25}$ (0.63M) & 86.88$_{\pm 0.46}$ (0.61M) \\
& DoRA       & 38.37$_{\pm 0.22}$ (2.38M) & 54.12$_{\pm 0.00}$ (2.45M) & 85.34$_{\pm 0.16}$ (0.69M) & 87.14$_{\pm 0.50}$ (0.65M) \\
& ProDiaL    &  \textbf{38.63}$_{\pm 0.25}$ (2.42M) & \textbf{55.09}$_{\pm 0.26}$ (2.38M) & \textbf{85.48}$_{\pm 0.13}$ (0.67M) & \textbf{87.91}$_{\pm 0.24}$ (0.65M) \\  \hline
\end{tabular}
\caption{\textbf{Multiple Runs for Both Projectors.} Average performance across three runs with random seeds 30, 42, and 100.}
\label{table:multi-run}
\end{table*}

\begin{table}[!t]
\small
\centering
\begin{tabular}{c|ccccc}
\hline
 Learning rate & 1e-5 & 5e-5 & \textbf{1e-4} & 5e-4 & 1e-3 \\ \hline
 LoRA & 36.15 & 37.86 & \textbf{38.33} & 38.31 & 37.47 \\
 DoRA & 36.15 & 37.66 & \textbf{38.13} & 37.33 & 36.09 \\ 
 ProDiaL & 36.17 & 37.8 & \textbf{38.92} & 37.64 & 36.46 \\ \hline 
\end{tabular}
\caption{\textbf{Optimal Learning Rates for LoRA, DoRA, and ProDiaL.} This experiment is conducted on HellaSwag dataset.}
\label{tab:learning rate}
\end{table}

\section{Experiment Details}
\label{sec:experiment_details}

\subsection{Models \& Datasets}

First, Mamba LLM~\cite{gu2023mamba} is the first model to implement the Mamba architecture, achieving faster inference than transformer-based LLMs as input token sizes increase. For Mamba LLM, we adapt the model pretrained on the PILE dataset~\cite{gao2020pile} to other reasoning task datasets: HellaSwag~\cite{zellers2019hellaswag}, Winogrande~\cite{sakaguchi2021winogrande}, ARC-Easy~\cite{clark2018think}, and ARC-Challenge~\cite{clark2018think}. The HellaSwag dataset is a challenging benchmark for commonsense reasoning that requires contextual understanding to predict the most plausible continuation of a given scenario from multiple choices. The Winogrande dataset is a large-scale benchmark for commonsense reasoning, consisting of sentence pairs with subtle differences, requiring the model to determine the best sentence completion by resolving nuanced context clues and pronoun references. The ARC-Easy dataset, a subset of the AI2 Reasoning Challenge (ARC), contains straightforward science questions at elementary and middle school levels, designed to assess a model’s basic factual and scientific reasoning abilities. The ARC-Challenge dataset, also part of the AI2 Reasoning Challenge, includes complex science questions that require advanced reasoning and domain knowledge.

Next, similar that transformer architecture was originally designed for language models but being used for vision tasks~\cite{dosovitskiy2020image}, the Mamba architecture, initially designed for language models, also has been adapted into Vision Mamba~\cite{zhu2024vision} by sequentially processing image tokens using both forward and backward State Space Models (SSMs)~\cite{gu2021efficiently}. For the Vision Mamba model, we adapt the model pretrained on the ImageNet dataset~\cite{deng2009imagenet} to other classification datasets, including Stanford Cars~\cite{krause20133d}, Caltech~\cite{li_andreeto_ranzato_perona_2022}, and Flowers~\cite{nilsback2008automated}. The Stanford Cars dataset includes images of 196 car models spanning various makes and years, offering detailed visual information to support the development of models capable of distinguishing between different car types and designs. The Caltech101 dataset comprises images from 101 object categories with diverse shapes and appearances, providing a foundation for developing models capable of recognizing real-world objects. The Flowers102 dataset contains images of 102 different flower species, capturing a range of visual variations to help models learn fine-grained distinctions among flower types.

\begin{table*}[!t]
\centering
\begin{tabular}{c|c|cccc}
\hline
 Method & Settings      & HellaSwag & Winogrande & ARC-E & ARC-C \\ \hline
\multirow{3}{*}{Default} 
              & Learning Rate &   1e-4    &      5e-6  &  5e-6 & 1e-5      \\ 
        & Total Training Iter (M) & 300K &    100K    &  100K & 100K      \\ 
        & Sampling Period (N) &  10K     &    5K      &   5K  & 5K      \\ \hline
\multirow{4}{*}{ProDiaL} 
& Both-Proj $(r_{b1}, r_{b2})$ & (768, 1536) & (768, 1536) & (64, 64) & (64, 64) \\ 
 & In-Proj $r_{b1}$           & 768      &    768     &  32   & 32      \\ 
 & Out-Proj $r_{b2}$          &  1536    &  1536      &  128  & 128      \\ 
 & Off-diagonal $r_\epsilon$   &   16     &     8      &  8    & 8     \\ \hline
\multirow{2}{*}{LoRA} & Low Rank r & 16 &   8        &  16   & 16     \\ 
          & Scaling factor $\alpha$ & 16 &   8        &  16   & 16    \\ \hline
\multirow{2}{*}{DoRA} & Low Rank r & 16 &   8        &  16   & 16    \\ 
          & Scaling factor $\alpha$ & 16 &   8        &  16   & 16    \\ \hline
\end{tabular}
\caption{\textbf{Fine-Tuning Settings for Mamba-130M.} This table summarizes the hyperparameters and configurations used for training and fine-tuning the Mamba-130M model across reasoning tasks (HellaSwag, Winogrande, ARC-E, and ARC-C). The Default settings include learning rates, total training iterations (in millions), and checkpoint sampling periods ($N$). For ProDiaL, specific configurations for the block-diagonal matrix ($r_{b1}, r_{b2}$) in input and output projectors, as well as the low-rank value ($r_\epsilon$) for off-diagonal matrices, are provided. Baseline methods (LoRA and DoRA) use a consistent low-rank value ($r$) and scaling factor ($\alpha$) for comparison.}

\label{tab:Mamba-130M hyperparameters}
\end{table*}

\begin{table*}[!t]
\centering
\begin{tabular}{c|c|cc}
\hline
Method & Settings      & Mamba-370M & Mamba-1.4B \\ \hline
\multirow{3}{*}{Default} 
              & Learning Rate &   5e-7    &      5e-7        \\ 
& Total Training Iter (M) & 30K &    30K          \\ 
& Sampling Period (N) &  1K     &    1K            \\ \hline
\multirow{4}{*}{ProDiaL} 
& Both-Proj $(r_{b1}, r_{b2})$ & (1024, 2048) & (2048, 4096)  \\ 
& In-Proj $r_{b1}$           & 1024      &    2048          \\ 
& Out-Proj $r_{b2}$          &  2048    &  4096            \\ 
& Off-diagonal $r_\epsilon$   &   16     &     64           \\ \hline
\multirow{2}{*}{LoRA} & Low Rank r & 16 &   64             \\ 
          & Scaling factor $\alpha$ & 16 &   64           \\ \hline
\multirow{2}{*}{DoRA} & Low Rank r & 16 &   64           \\ 
         & Scaling factor $\alpha$ & 16 &   64            \\ \hline
\end{tabular}
\caption{\textbf{Fine-Tuning Settings for Larger Mamba Models.} This table outlines the hyperparameters and configurations used for training and fine-tuning Mamba-370M and Mamba-1.4B on the Winogrande dataset.}

\label{tab:Mamba Large hyperparameters}
\end{table*}

\begin{table}[!t]
\centering
\small
\begin{tabular}{c|c|ccc}
\hline
 & Settings & StanfordCars & Caltech & Flowers \\ \hline
\multirow{4}{*}{{\centering\footnotesize\textbf{ProDiaL}}} & ($r_{b1}$, $r_{b2}$) & (192,384) & (192,384) & (16,16) \\
& $r_{b1}$ & 192 & 192 & 8 \\
& $r_{b2}$ & 384 & 384 & 32 \\
& $r_\epsilon$ & 16 & 16 & 8 \\ \hline
\multirow{2}{*}{{\centering\footnotesize\textbf{LoRA}}} & $r$ & 16 & 16 & 16 \\
& $\alpha$ & 16 & 16 & 16 \\ \hline
\multirow{2}{*}{{\centering\footnotesize\textbf{DoRA}}} & $r$ & 16 & 16 & 16 \\
& $\alpha$ & 16 & 16 & 16 \\ \hline
\end{tabular}
\caption{\textbf{Hyperparameters for Fine-Tuning Vim-tiny on Classification Tasks.} The table presents the hyperparameters used for fine-tuning Vim-tiny on the StanfordCars, Caltech, and Flowers datasets. $r_{b1}$ and $r_{b2}$ represent the low ranks for block-diagonal matrices in input and output projectors, respectively. For LoRA and DoRA methods, $r$ indicates the low rank, while $\alpha$ denotes the scaling factor.}

\label{tab:Vim-tiny hyperparameters}
\end{table}

\begin{table*}[!t]
\centering
\begin{tabular}{c|c|cccc}
\hline
 Method & Settings      & HellaSwag & Winogrande & ARC-E & ARC-C \\ \hline
\multirow{3}{*}{Default} 
              & Learning Rate &   1e-4    &      1e-6  &  1e-5 & 5e-6      \\ 
        & Total Training Iter (M) & 200K &    30K    &  100K & 100K      \\ 
        & Sampling Period (N) &  10K     &    1K      &   5K  & 5K      \\ \hline
\multirow{4}{*}{ProDiaL} 
& Both-Proj $(r_{b1}, r_{b2})$ & (64, 64) & (64, 128) & (128, 64) & (32, 128) \\ 
 & In-Proj $r_{b1}$           & 32      &    768     &  128   & 256      \\ 
 & Out-Proj $r_{b2}$          &  128    &  1536      &  256  & 128      \\ 
 & Off-diagonal $r_\epsilon$   &   8     &     8      &  8    & 8     \\ \hline
\multirow{2}{*}{LoRA} & Low Rank r & 16 &   16        &  16   & 16     \\ 
          & Scaling factor $\alpha$ & 16 &   16        &  16   & 16    \\ \hline
\multirow{2}{*}{DoRA} & Low Rank r & 16 &   16        &  16   & 16    \\ 
          & Scaling factor $\alpha$ & 16 &   16        &  16   & 16    \\ \hline
\end{tabular}
\caption{\textbf{Fine-Tuning Settings for Mamba2-130M.} This table details the hyperparameters and configurations used for fine-tuning Mamba2-130M on reasoning tasks, including learning rates, training steps, checkpoint intervals, and settings for ProDiaL, LoRA, and DoRA methods.}

\label{tab:Mamba2-130M hyperparameters}
\end{table*}

\subsection{Universal Effectiveness of Experiments}

All experiments and conclusions, including the effectiveness of fine-tuning only the projectors and our proposed ProDiaL approach, are universally applicable. This universal applicability has been empirically validated through multiple trials conducted under various random seeds. As demonstrated in \cref{table:multi-run}, the performance remains consistently robust across various random seeds, reinforcing the reliability of our method. This consistency highlights the stability of our approach, suggesting that it is not highly dependent on specific initialization conditions or random factors. 

\subsection{Optimal Learning Rates Selection for Fair Comparison}

To ensure a fair comparison between LoRA, DoRA, and our proposed ProDiaL method, we conducted a comprehensive search to identify optimal learning rates for each approach.  As shown in~\cref{tab:learning rate}, LoRA, DoRA, and ProDiaL achieve their highest performance at the \textbf{same learning rate}. We believe this convergence is due to the inherent design similarities between the methods, as both DoRA and ProDiaL build upon the LoRA framework. Consequently, the reported results are based on these optimized settings to ensure that each method performs at its best.

\subsection{Training and Evaluation}
\label{sec:training and evaluation}

Our experiments in main manuscript are mainly conducted on Vim~\cite{zhu2024vision} based on Mamba 2 architecture~\cite{dao2024transformers} and Mamba LLM~\cite{gu2023mamba} based on Mamba 1 architecture. Below, we detail the experimental settings for each dataset.

\subsubsection{Mamba LLM Experiments}

In the experiments presented in Table 3 of the main manuscript, we use the Mamba-130M, configured with 24 layers and a maximum sequence length of 512. The dimensions for the input projectors are set as follows: input projectors have $d_{in} = 768$ and $d_{out} = 3072$, while output projectors are configured with $d_{in} = 1536$ and $d_{out} = 768$.

For the results in Table 4 of the main manuscript, we employ the larger Mamba-370M and Mamba-1.4B models, both configured with 48 layers and a maximum sequence length of 512. Mamba-370M uses input projectors with $d_{in} = 1024$ and $d_{out} = 4096$ and output projectors with $d_{in} = 2048$ and $d_{out} = 1024$. For Mamba-1.4B, the input projectors have $d_{in} = 2048$ and $d_{out} = 8192$, while the output projectors have $d_{in} = 4096$ and $d_{out} = 2048$.

For training Mamba LLM models, we use the AdamW optimizer with a batch size of 4 and a constant learning rate scheduler. Additional hyperparameters including ProDiaL's settings, which control the number of learnable parameters, are provided in~\cref{tab:Mamba-130M hyperparameters,tab:Mamba Large hyperparameters}.

For evaluation, we use Language Model Evaluation Harness framework\footnote[1]{https://github.com/EleutherAI/lm-evaluation-harness}, following~\cite{gu2023mamba}\footnote[2]{https://github.com/state-spaces/mamba}.

\subsubsection{Vision Mamba Experiments}

In the experiments in Table 3 of the main manuscript, we use the Vim-tiny, which has 24 layers, a patch size of 16, and an input image size of 224. The input projectors are set with $d_{in} = 192$ and $d_{out} = 768$, and output projectors have $d_{in} = 384$ and $d_{out} = 192$. We train the Vim-tiny model for 300 epochs with a batch size of 128, using the AdamW optimizer with a learning rate of $5e-4$ and weight decay of $0.1$. A cosine learning rate scheduler with 5 warm-up epochs starting from $1e-6$ is applied. Hyperparameters for ProDiaL, determining the number of learnable parameters, are detailed in~\cref{tab:Vim-tiny hyperparameters}.

In Table 4 of the main manuscript, we use the Vim-small, which also has 24 layers, a patch size of 16, and an input image size of 224. Whereas, input projectors are configured with $d_{in} = 384$ and $d_{out} = 1536$, while output projectors have $d_{in} = 768$ and $d_{out} = 384$. Training for Vim-small spans 300 epochs with a batch size of 64, using the AdamW optimizer at a learning rate of $1e-3$, weight decay of $0.05$, and dropout rate of $0.05$. The cosine learning rate scheduler is also used with a 5-epoch warm-up starting from $1e-6$. For the Caltech dataset, ProDiaL hyperparameters are set as follows: $r_{b1} = 192$, $r_{b2} = 384$, and $r_\epsilon = 16$.

\subsection{Details for Each Analysis}

In this section, we describe the detailed settings for each analysis experiment in the tables and figures of the main manuscript. The training and evaluation details for all models and datasets follow the guidelines provided in Section~\ref{sec:training and evaluation}

\textbf{Table 1} and \textbf{Table 2} in main manuscript present the ablation studies of Vision Mamba and Mamba LLM, respectively. For these experiments, the components of the Vim-tiny model were trained on the SUN dataset~\cite{xiao2010sun}, and the components of the Mamba-130M model were trained on the HellaSwag dataset. The results from both tables consistently show that the Projectors contribute more significantly to capturing knowledge for downstream tasks compared to the State-Space Model (SSM).

\textbf{Figure 2} in main manuscript visualizes the importance of training diagonal entries in the linear transformation matrix $T$ for learning an effective linear transformation. This analysis was conducted using the Vim-tiny model trained on the Caltech dataset. The visualization illustrates the input projector at the fourth layer, which was randomly selected. Similar characteristics were observed in other layers and the output projectors, reinforcing the generalizability of this observation.

\textbf{Figure 4} in main manuscript empirically demonstrates the effectiveness of training diagonal entries in the linear transformation matrix $T$. This experiment, conducted on the Vim-tiny model with the Caltech dataset, explores four distinct configurations for fine-tuning $T$. In the first configuration, the entire $T$ matrix is directly fine-tuned, initialized as an identity matrix to ensure it starts as a standard linear transformation. The second configuration fine-tunes a block-diagonal matrix, where each block is initialized as an identity matrix.  In the third configuration, only a diagonal vector was fine-tuned, initialized as a vector of ones. The fourth configuration involved fine-tuning only the off-diagonal matrix, starting from a zero matrix with the diagonal entries masked.

\textbf{Table 5} in main manuscript investigates whether the diagonal entries in the linear transformation matrix $T$ are effective only for Projectors or across all linear layers in the Mamba architecture. This experiment was also conducted using the Vim-tiny model on the Caltech dataset.

\textbf{Table 6} in main manuscript explores the role of non-attention modules in the Transformer architecture. For this analysis, the Vision Transformer (ViT-B/16) model pretrained on ImageNet was fine-tuned on the CIFAR-100 dataset. The model was trained using the Adam optimizer with a batch size of 128 and a learning rate of 0.001 for 5000 iterations. A cosine learning rate scheduler with a warmup period of 500 steps was used. For LoRA adaptation, a low rank of 8 was applied to both the Attention and FFN modules.

\textbf{Table 7} in main manuscript provides an ablation studies of ProDiaL components, including Block-diagonal matrix $D_b$, Off-diagonal matrix $\epsilon$, and scaling factor $s$. This experiment was conducted on the Vim-tiny model fine-tuned on the Caltech dataset.

\end{document}